%% file: main.tex
\documentclass[acmsmall]{acmart}

\usepackage{booktabs} 
\usepackage{xcolor}
\usepackage{graphicx}
\usepackage{amsmath}

\usepackage{subfigure}
\usepackage[ruled,vlined]{algorithm2e}
\SetKwInput{KwInput}{Input}                
\SetKwInput{KwOutput}{Output}              
\SetKwBlock{spprocess}{Search Process:}{end}

\usepackage{titlesec}
\titlespacing\subsection{0pt}{3pt plus 1pt minus 1pt}{3pt plus 1pt minus 1pt}
\titlespacing\subsubsection{0pt}{3pt plus 1pt minus 1pt}{3pt plus 1pt minus 1pt}

\usepackage{mathtools}
\usepackage{multirow}
\usepackage{textcomp}

\AtBeginDocument{%
  \providecommand\BibTeX{{%
    \normalfont B\kern-0.5em{\scshape i\kern-0.25em b}\kern-0.8em\TeX}}}

\setcopyright{none}

\begin{document}

\title{FLASH:   \underline{F}ast Neura\underline{l} \underline{A}rchitecture \underline{S}earch  with \underline{H}ardware Optimization}
\author[1]{Guihong Li}
\email{lgh@utexas.edu}
\affiliation{%
  \institution{The University of Texas at Austin}
  \city{Austin}
  \state{Texas}
  \country{USA}
}
\author[2]{Sumit K. Mandal}
\email{skmandal@wisc.edu}
\affiliation{%
  \institution{University of Wisconsin–Madison}
  \city{Madison}
  \state{Wisconsin}
  \country{USA}
}

\author[2]{Umit Y. Ogras}
\email{uogras@wisc.edu}
\affiliation{%
  \institution{University of Wisconsin–Madison}
  \city{Madison}
  \state{Wisconsin}
  \country{USA}
}

\author{Radu Marculescu}
\email{radum@utexas.edu}
\affiliation{%
  \institution{The University of Texas at Austin}
  \city{Austin}
  \state{Texas}
  \country{USA}
}

\renewcommand{\shortauthors}{G. Li, et al.}

\begin{abstract}
Neural architecture search (NAS) is a promising technique to design efficient and high-performance deep neural networks (DNNs).
As the performance requirements of ML applications grow continuously, the hardware accelerators start playing a central role in DNN design. This trend makes NAS even more complicated and time-consuming for most real applications. This paper proposes FLASH, a very fast NAS methodology that co-optimizes the DNN accuracy and performance on a real hardware platform. 
As the main theoretical contribution, we first propose the NN-Degree, an analytical metric to quantify the topological characteristics of DNNs with skip connections (e.g., DenseNets, ResNets, Wide-ResNets, and MobileNets). The newly proposed NN-Degree allows us to do \textit{training-free} NAS within one second and build an accuracy predictor by training as few as 25 samples out of a vast search space with more than 63 billion configurations. Second, by performing inference on the target hardware, we fine-tune and validate our analytical models to estimate the latency, area, and energy consumption of various DNN architectures while executing standard ML datasets. 
Third, we construct a hierarchical algorithm based on simplicial homology global optimization (SHGO) to optimize the model-architecture co-design process, while considering the area, latency, and energy consumption of the target hardware.
We demonstrate that, compared to the state-of-the-art NAS approaches, our proposed hierarchical SHGO-based algorithm enables more than four orders of magnitude speedup (specifically, the execution time of the proposed algorithm is about 0.1 seconds). 
Finally, our experimental evaluations show that FLASH is easily transferable to different hardware architectures, thus enabling us to do NAS on a Raspberry Pi-3B processor in less than 3 seconds. 
\end{abstract}

\ccsdesc[500]{Computing methodologies~Artificial intelligence}
\ccsdesc[500]{Computing methodologies~Computer vision}
\ccsdesc[500]{Computer systems organization~Embedded systems}

\keywords{Neural Networks, Network Science, Hardware Optimization, Neural Architecture Search, Model-Architecture Co-design, Resource-constrained Devices}

\maketitle

\input{0-abstract.tex}
\input{1-introduction.tex}
\input{2-related_work.tex}
\input{3-approach.tex}

\input{4-experimental_results.tex}

\input{5-conclusion.tex}

\section{Acknowledgments}
This work was supported in part by the US National Science Foundation (NSF) grant CNS-2007284, and in part by Semiconductor Research Corporation (SRC) grants GRC 2939.001 and 3012.001.

\bibliographystyle{ACM-Reference-Format}

\bibliography{acmart}

\end{document}

%% file: 0-abstract.tex




%% file: 1-introduction.tex
\section{Introduction} 
\label{sec:introduction}
During the past decade, deep learning (DL) has led to significant breakthroughs in many areas, such as image classification and natural language processing~\cite{densenet,resnet,big_nlp}. However, the existing large models and computation complexity limit the deployment of DL on resource-constrained devices and its large-scale adoption in edge computing. Multiple model compression techniques, such as network pruning~\cite{han_prune}, quantization~\cite{bnn}, and knowledge distillation~\cite{kd}, have been proposed to compress and deploy such complex models on resource-constrained devices without sacrificing the test accuracy.
However, these techniques require a significant amount of manual tuning. Hence, neural architecture search (NAS) has been proposed to automatically design neural architectures with reduced model sizes~\cite{baker_17, quoc_le, lstm, Darts,elsken2019neural}.

NAS is an optimization problem with specific targets (e.g., high classification accuracy) over a set of possible candidate architectures. The set of candidate architectures defines the (typically vast) search space, while the optimizer defines the search algorithm. Recent breakthroughs in NAS can simplify the tricky (and error-prone) ad-hoc architecture design process~\cite{lstm, hyper_nas}. Moreover, the networks obtained via NAS have higher test accuracy and significantly fewer parameters than the hand-designed networks~\cite{Darts, real_17}. These advantages of NAS have attracted significant attention from researchers and engineers alike~\cite{nas_survey}. However, most of the existing NAS approaches do not explicitly consider the hardware constraints (e.g., latency and energy consumption). Consequently, the resulting neural networks still cannot be deployed on real devices. 

To address this drawback, recent studies propose \textit{hardware-aware NAS}, which incorporates the hardware constraints of networks during the search process~\cite{jiang2020device}. Nevertheless, current approaches are time-consuming since they involve training the candidate network, and a tedious search process~\cite{wu2019fbnet}.
To accelerate NAS, recent NAS approaches rely on graph neural networks (GNNs) to estimate the accuracy of a given network~\cite{eccv_gates, yiran_gnn, brp_nas, pr_2020_gnn_acc_pre}. However, training a GNN-based accuracy predictor is still time-consuming (in the order of tens of minutes~\cite{chiang2019cluster} to hours~\cite{mao2019learning} on GPU clusters). 
Therefore, adapting existing NAS approaches to different hardware architecture is challenging due to their intensive computation and execution time requirements.

To alleviate the computation cost of current NAS approaches, we propose to analyze the NAS problem from a \textit{network topology} perspective. This idea is motivated by observing that the tediousness and complexity of current NAS approaches stem from the lack of understanding of what actually contributes to a neural network's accuracy. Indeed, the innovations on the topology of neural architecture, especially the introduction of skip connections, have achieved great success in many applications~\cite{densenet,resnet}. This is because, in general, the network topology (or structure) strongly influences the phenomena taking place over them~\cite{newman2006structure}. For instance, how closely the social network users are interconnected directly affects how fast the information propagates through the network~\cite{barabasi2003scale_Free}. 
Similarly, a DNN architecture can be seen as a network of connected neurons. As discussed in~\cite{nn_mass}, the topology of deep networks has a significant impact on how effectively the gradients can propagate through the network and thus the test performance of neural networks. These observations motivate us to take an approach from network science to quantify the topological property of neural networks to accelerate NAS.

From an application perspective, the performance and energy efficiency of DNN accelerators are other critical metrics besides the test accuracy. In-memory computing (IMC)-based architectures have recently emerged as a promising technique to construct high-performance and energy-efficient hardware accelerators for DNNs. IMC-based architectures can store all the weights on-chip, hence removing the latency occurring from off-chip memory accesses. However, IMC-based architectures face the challenge of a tremendous increase of on-chip communication volume. While most of the state-of-the-art neural networks adopt skip connections in order to improve their performance~\cite{resnet, mobilenetv2, densenet}, the wide usage of skip connections requires large amounts of data transfer across multiple layers, thus causing a significant communication overhead. Prior work on IMC-based DNN accelerators proposed bus-based network-on-chip (NoC)~\cite{chen2018neurosim} or cmesh-based NoC~\cite{shafiee2016isaac} for communication between multiple layers. However, both bus-based and cmesh-based on-chip communication significantly increase the area, latency, and energy consumption of hardware; hence, they do not offer a promising solution for future accelerators.

Starting from these overarching ideas, this paper proposes FLASH -- a fast neural architecture search with hardware optimization -- to address the drawbacks of current NAS techniques. FLASH delivers a neural architecture that is co-optimized with respect to accuracy and hardware performance. Specifically, by analyzing the topological property of neural architectures from a network science perspective, we propose a new topology-based metric, namely, the \textit{NN-Degree}. We show that NN-Degree could indicate the test performance of a given architectures. This makes our proposed NAS \textit{training-free} during the search process and accelerates NAS by orders of magnitude compared to state-of-the-art approaches. Then, we demonstrate that NN-Degree enables a lightweight accuracy predictor with only \textit{three parameters}. Moreover, to improve the on-chip communication efficiency, we adopt the mesh-NoC for the IMC-based hardware. Based on the communication-optimized hardware architecture, we measure the hardware performance for a subset of neural networks from the NAS search space. 
Then, we construct analytical models for the area, latency, and energy consumption of a neural network based on our optimized target hardware platform.
Unlike existing neural network-based and black-box style searching algorithms~\cite{jiang2020device}, the proposed NAS methodology enable searching across the entire search space via a mathematically rigorous and time-efficient optimization algorithm.
Consequently, our experimental evaluations show that FLASH significantly pushes forward the NAS frontier by enabling NAS in less than 0.1 seconds on a 20-core Intel Xeon CPU. Finally, we demonstrate that FLASH could be readily transferred to other hardware platforms (e.g., Raspberry Pi) only by fine-tuning the hardware performance models.

Overall, this paper makes the following contributions:
%
\begin{itemize}

    \item We propose a new topology-based analytical metric (\textit{NN-Degree}) to quantify the topological characteristics of DNNs with skip connections. We demonstrate that the NN-Degree enables a \textit{training-free} NAS within seconds. Moreover, we use the NN-Degree metric to build a new lightweight (\textit{three-parameter}) accuracy predictor by training as few as 25 samples out of a vast search space with more than 63 billion configurations.
    Without any significant loss in accuracy, our proposed accuracy predictor requires 6.88$\times$ fewer samples and provides a $65.79\times$ reduction of the fine-tuning time cost compared to existing GNN/GCN based approaches~\cite{yiran_gnn}. 
    \item We construct analytical models to estimate the latency, area, and energy consumption of various DNN architectures. We show that our proposed analytical models are applicable to multiple hardware architectures and achieve a high accuracy with less than one second fine-tuning time cost.
    \item We design a hierarchical simplicial homology global optimization (SHGO)-based algorithm, to search for the optimal architecture. Our proposed hierarchical SHGO-based algorithm enables 27729$\times$ faster (less than 0.1 seconds) NAS compared to RL-based baseline approach.
    \item We demonstrate that our methodology enables NAS on a Raspberry Pi 3B with less than 3 seconds computational time. To our best knowledge, this is the first work showing NAS running directly on edge devices with such low computational requirements.
\end{itemize}

The rest of the paper is organized as follows. In Section \ref{sec:related_work},
we discuss related work and background information. 
In Section \ref{sec:methodology}, we formulate the optimization problem, then describe the new analytical models and search algorithm. Our experimental results are presented in Section \ref{sec:experimental_results}. Finally, Section \ref{sec:conclusion} concludes the paper with remarks on our main contributions and future research directions.

%% file: 2-related_work.tex
\section{Related Work and Background Information} 
\label{sec:related_work}


\noindent\textbf{Hardware-aware NAS:} Hardware accelerators for DNNs have recently become popular due to high-performance demand for multiple applications~\cite{img_net, manning1999foundations,benmeziane2021comprehensive};
they can reduce the latency and energy associated with DNN inference significantly.
The hardware performance (e.g., latency, energy, and area) of accelerators varies with DNN properties (e.g., number of layers, parameters, etc.);
therefore, hardware performance also is a crucial factor to consider during NAS.

Several recent studies consider hardware performance for NAS. Authors in~\cite{jha_dac20} introduce a growing and pruning strategy that automatically maximizes the test accuracy and minimizes the FLOPs of neural architectures during training. 
A platform-aware NAS targeting mobile devices is proposed in~\cite{global_1};
the objective is to maximize the model accuracy with an upper bound on latency. Authors in~\cite{wu2019fbnet} create a latency-aware loss function to perform differentiable NAS. The latency of DNNs is estimated through a lookup table which consists of the latency of each operation/layer.
However, both of these studies consider latency as the only metric for hardware performance. Authors in~\cite{diana_modeling} propose a hardware-aware NAS framework to design convolutional neural networks. Specifically, by building analytical latency, power, and memory models, they create a hardware-aware optimization methodology to search for the optimal architecture that meets the hardware budgets. 
Authors in~\cite{jiang2020device} consider latency, energy, and area as metrics for hardware performance while performing NAS.
Also, a reinforcement learning (RL)-based controller is adopted to tune the network architecture and device parameters.
The resulting network is retrained to evaluate the model accuracy.
There are two major drawbacks of this approach.
First, RL is a slow-converging process that prohibits fast exploration of the design space.
Second, retraining the network further exacerbates the search time leading to hundreds of GPU hours needed for real applications~\cite{quoc_le}. 
Furthermore, most existing hardware-aware NAS approaches explicitly optimize the architectures for a specific hardware platform~\cite{cai2018proxylessnas, wu2019fbnet, edd_Dac}. Hence, if we switch to some new hardware, we need to repeat the entire NAS process, which is very time-consuming under the existing NAS frameworks~\cite{cai2018proxylessnas, wu2019fbnet, edd_Dac}. The demand for reducing the overhead of adaptation to new hardware motivates us to improve the transferability of hardware-aware NAS methodology. 

\noindent\textbf{Accuracy Predictor-based NAS:} Several approaches perform NAS by estimating the accuracy of the network~\cite{eccv_gates,yiran_gnn,brp_nas,pr_2020_gnn_acc_pre}.
These approaches first train a graph neural network (GNN), or a graph convolution network (GCN), to estimate the network accuracy while exploring the search space. During the searching process, the test accuracy of the sample networks is obtained from the estimator instead of doing regular training.
Although by estimating the accuracy, the NAS process is significantly accelerated, the training cost of the accuracy predictor itself remains a bottleneck.
GNN requires many training samples to achieve high accuracy, thus involving a significant overhead during training the candidate networks from the search space.
Therefore, using accuracy predictors to do NAS still suffers from excessive computation and time requirements.

\noindent\textbf{Time-efficient NAS:} To reduce the time cost of training candidate networks, authors in~\cite{hyper_nas, single_path_nas} introduced the weight sharing mechanism (WS-NAS). Specifically, candidate networks are generated by randomly sampling part of a large network (supernet). Hence, candidate networks share the weights of the supernet and update these weights during training. By reusing these trained weights instead of training from scratch, WS-NAS significantly improves the time efficiency of NAS. However, the accuracy of these models obtained via WS-NAS is typically far below those obtained from training from scratch. Several optimization techniques have been proposed to fill the accuracy gap between the shared weights and stand-alone training~\cite{big_nas, Cai2020Once-for-All}. For example, authors in \cite{Cai2020Once-for-All} propose a progressive shrinking algorithm to train the supernet. However, in many cases, the resulting networks still need some fine-tuning epochs to get the final architecture. To further accelerate the NAS process, some works propose the differentiable NAS to accelerate the NAS process~\cite{Darts,cai2018proxylessnas}. The differentiable NAS approaches search for the optimal architecture by learning the optimal architecture parameters during the training process. Hence, differentiable NAS only needs to train the supernet once, thus reducing the training time significantly. Nevertheless, due to the significantly large number of parameters of the supernet, differentiable NAS requires a high volume of GPU memory. In order to further improve the time-efficiency of NAS, several approaches have been proposed to do training-free NAS~\cite{tf_nas1,tf_nas2}. These approaches leverage some training-free proxy that indicates the test performance of some given architectures; hence, the training time is eliminated from the entire NAS process. However, these methods usually use gradient-based information to build the proxy~\cite{tf_nas1,tf_nas2}. Therefore, in order to calculate the gradients, GPUs are still necessary for the backward propagation process. To totally decouple the NAS process from using GPU platforms, our work proposes a GPU-free proxy to do training-free NAS. We provide more details in Section~\ref{subsec:tf_nas}.

\begin{figure} [b]
    \centering
    \includegraphics[width=0.8\textwidth]{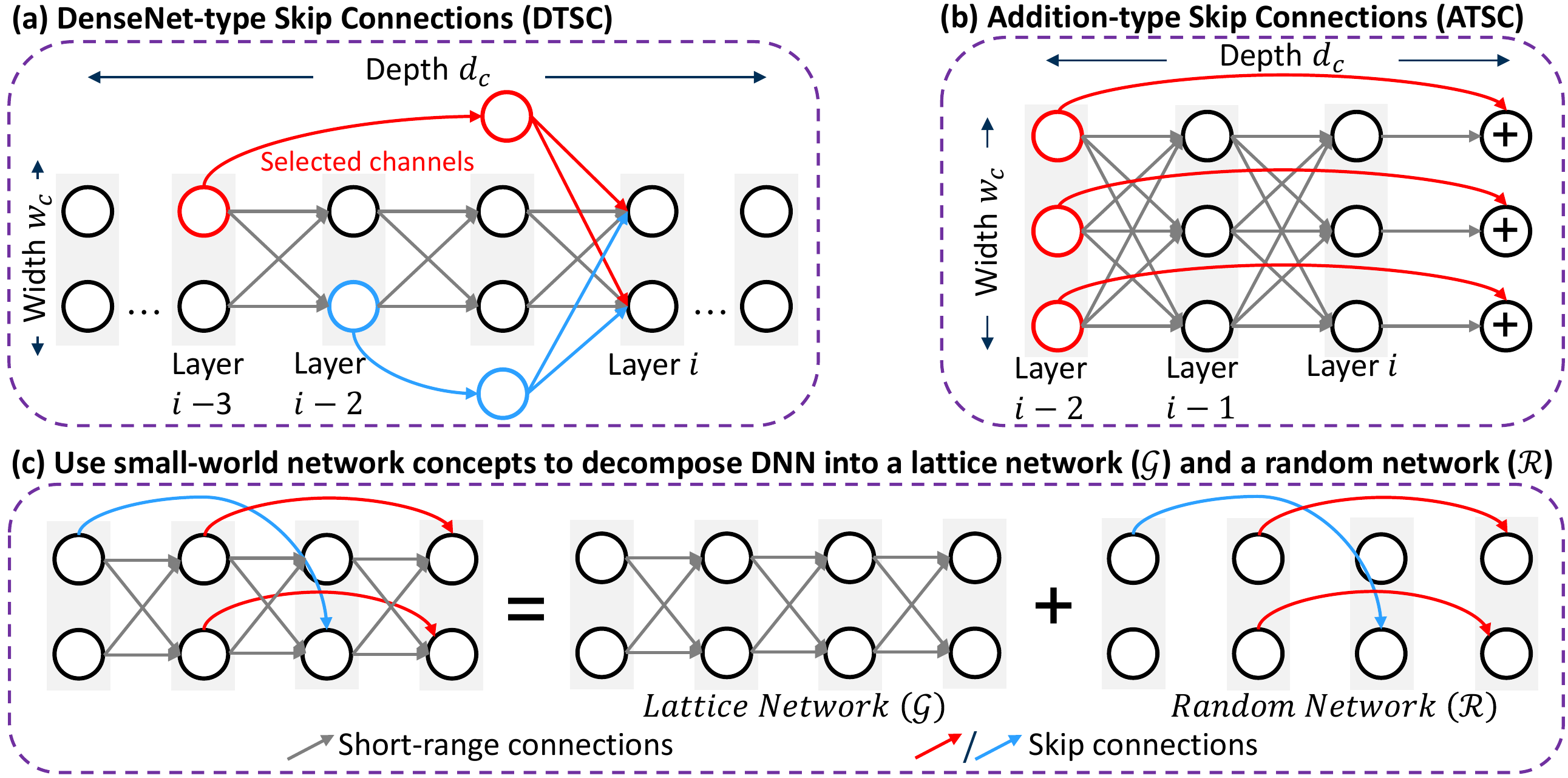}
    \caption{Modeling a CNN as a network in network science: Each channel is modeled as a node; each convolution kernel/filter is modeled as a link/connection. (a) Illustration of a single cell with DenseNet-type skip connections (DTSC). (b) Illustration of a single cell with Addition-type skip connections (ATSC). (c) Decomposition of a network cell with skip connections into a Lattice Network $\mathcal{G}$ and a Random Network $\mathcal{R}$.}
    \label{fig:skip_link}
\end{figure}

\noindent\textbf{Skip connections and Network Science:} Currently, both networks obtained by manual design and NAS have shown that long-range links (i.e., skip connections) are crucial for getting higher accuracy~\cite{resnet,densenet,mobilenetv2,Darts}. Overall, there are two commonly used skip connections in neural networks. First, we have the \textit{DenseNet-type} skip connections (DTSC), which concatenate previous layers' outputs as the input for the next layer~\cite{densenet}. To study the topological properties and enlarge the search space, we do \textit{not} use the original DesneNets~\cite{densenet}, which contains all-to-all connections. Instead, we consider a generalized version where we vary the number of skip connections by randomly selecting only some channels for concatenation, as shown in Fig. \ref{fig:skip_link}(a). The other type of skip connections is the \textit{addition-type} skip connections (ATSC), which consist of links that bypass several layers to be directly added to the output of later layers (see Fig. \ref{fig:skip_link}(b))~\cite{resnet}. 

In network science, a small-world network is defined as a highly clustered network, thus showing a small distance (typically logarithmic in the number of network nodes) between any two nodes inside the network~\cite{smallworldness}.
Considering the skip connections in neural networks, 
we propose to use the \textit{small-world network} concept to analyze networks with both short- and long-range (or skip) links. Indeed, small-world networks can be decomposed into: (i) a lattice network $\mathcal{G}$ accounting for short-range links; (ii) a random network $\mathcal{R}$ accounting for long-range links (see Fig.~\ref{fig:skip_link}(c)).
The co-existence of a rich set of short- and long-range links leads to both a high degree of clustering and short average path length (logarithmic with network size).
We use the small-world network to model and analyze the topological property of neural networks in Section \ref{sec:methodology}.

\noindent\textbf{Average Degree:} The \textit{average degree} of a network determines the average number of connections a node has, i.e., the total number of edges divided by the total number of nodes. The average degree and degree distribution (i.e., distribution of node degree) are important topological characteristics that directly affect how information flows through a network \cite{barabasi2003scale_Free}. 
Indeed, the small network theory reveals that the average degree of a network has a significant impact on network average path length and clustering behavior \cite{smallworldness}. Therefore, we investigate the performance gains due to the topological properties by using network science.

%% file: 3-approach.tex
\section{Proposed Methodology} 
\label{sec:methodology}

\begin{figure}[t]
	\centering
	\includegraphics[width=0.8\textwidth]{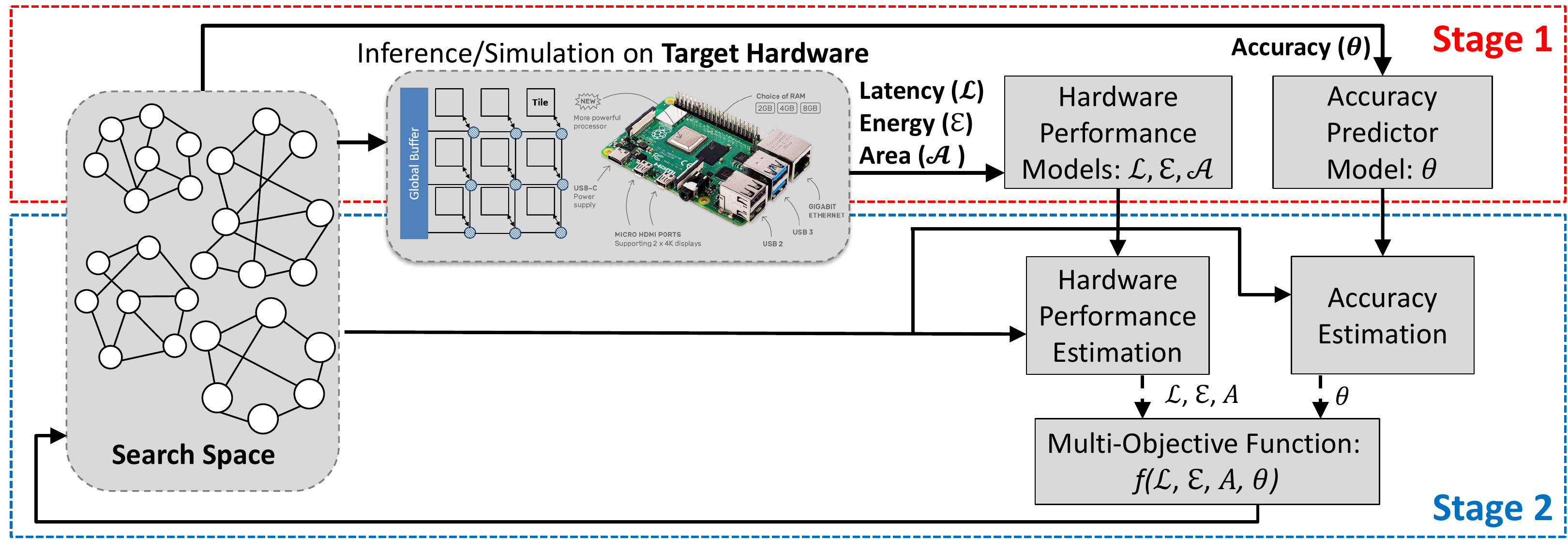}
    \caption{Overview of the proposed approach. Stage 1 (red box): we build hardware performance model (i.e., latency $\mathcal{L}$, energy $\mathcal{E}$, and area $\mathcal{A}$) and accuracy predictor by randomly sampling candidate networks from the search space to evaluate the hardware characteristics (latency $\mathcal{L}$, energy $\mathcal{E}$, and area $\mathcal{A}$) and test accuracy $\theta$. Stage 2 (blue box): search for the optimal network architecture given the multi-objective function $f(\mathcal{L}, \mathcal{E}, \mathcal{A}, \theta)$.}
	\label{fig:overview}
\end{figure}

\subsection{Overview of New NAS Approach}
The proposed NAS framework is a two-stage process, as illustrated in Fig. \ref{fig:overview}: 
(i) We first quantify the topological characteristics of neural networks by the newly proposed NN-Degree metric. Then, we randomly select a few networks and train them to fine-tune the accuracy predictor based on the network topology. We also build analytical models to estimate the latency, energy, and area of given neural architectures.
(ii) Based on the accuracy predictor and analytical performance models in the first stage, we use a simplical homology global optimization (\textit{SHGO})-based algorithm in a hierarchical fashion to search for the optimal network architecture.

\subsection{Problem Formulation of hardware-aware NAS}

The overall target of the hardware-aware NAS approach is to find the network architecture that gives the highest test accuracy while achieving small area, low latency, and low energy consumption when deployed on the target hardware. 
In practice, there are constraints (budgets) on the hardware performance and test accuracy. For example, battery-based devices have very constrained energy capacity~\cite{wang2020neural}. Hence, there is an upper bound for the energy consumption of the neural architecture. To summarize, the NAS problem can be expressed as:
\begin{equation}
\begin{aligned}
        \max
        &\quad f_{obj}=\frac{\theta}{\mathcal{A}\times\mathcal{L}\times\mathcal{E}} \\
     \text{subject\ to:} &\quad \mathcal{\theta} \geq \mathcal{\theta}_M,\ \mathcal{A} \leq \mathcal{A}_M,
     \ \mathcal{L} \leq \mathcal{L}_M,
     \ \mathcal{E} \leq \mathcal{E}_M\\
\end{aligned}
\label{eq:problem_definition}
\end{equation}

\noindent{where $\mathcal{\theta}_M$, $\mathcal{A}_M$, $\mathcal{L}_M$, and $\mathcal{E}_M$ are the constraints on the test accuracy, area, latency, and energy consumption, respectively. We summarize the symbols (and their meaning) in this part in Table \ref{table:prob_form}.}

\begin{table}[t]
\caption{Symbols and their corresponding definition/meaning used in our Problem Formulation.}
\scalebox{0.88}{\begin{tabular}
{|l|l|}
\hline
Symbol & Definition  \\
\hline
\hline
$f_{obj}$ & Objective function of NAS \\
\hline
$\theta$ & Test accuracy of a given network \\
\hline
 $A$ & Chip area \\
\hline
 $\mathcal{L}$ & Inference latency of a given network \\
\hline
 $\mathcal{E}$ & Inference energy consumption of a given network \\
\hline
$\theta_M$ & Constraint of test accuracy for NAS \\
\hline
 $A_M$ & Constraint of area for NAS \\
\hline
 $\mathcal{L}_M$ & Constraint of inference latency for NAS \\
\hline
 $\mathcal{E}_M$ & Constraint of inference energy consumption for NAS \\
\hline
\end{tabular}}
\label{table:prob_form}
\end{table}

\subsection{NN-Degree and Training-free NAS}


This section first introduces our idea of modeling a CNN based on network science~\cite{smallworldness}. To this end, we define a group of consecutive layers with the same width (i.e., number of output channels, $w_c$) as a \textit{cell}; then we break the entire network into multiple cells and denote the number of cells as $N_c$. Similar to MobileNet-v2~\cite{mobilenetv2}, we also adopt a width multiplier ($w_m$) to scale the width of each cell. 
Moreover, following most of the mainstream CNN architectures, we assume that each cell inside a CNN has the same number of layers ($d_c$). Furthermore, as shown in Fig.~\ref{fig:skip_link}, we consider each channel of the feature map as a node in a network and consider each convolution filter/kernel as an undirected link. These notations are summarized in Table \ref{table:acc_pred}.

\begin{table}[htb]
\caption{Symbols and their corresponding definition/meaning used in our NN-Degree based analytical accuracy predictor.}
\scalebox{0.88}{\begin{tabular}
{|l|l|}
\hline
Symbol & Definition  \\
\hline
\hline
$g$ & NN-Degree (new metric we propose) \\
 \hline
$g_\mathcal{G}$ &  NN-Degree of the lattice network (short-range connections)\\
 \hline
$g_\mathcal{R}$ &  NN-Degree of the random network (long-range or skip connections)\\
 \hline
$N_c$ & Number of cells \\
 \hline
$w_c$ & Number of output channels per layer within cell $c$ (i.e., the width of cell $c$)\\
\hline
$d_c$ & Number of layers within cell $c$ (i.e., the depth of cell $c$)\\
\hline
$SC_c$ & Number of skip connections within cell $c$   \\
\hline
$a_\theta,b_\theta,c_\theta$ & Learnable parameters for the accuracy predictor \\
\hline
\end{tabular}}
\label{table:acc_pred}
\end{table}

Combining the concept of small-world networks in Section \ref{sec:related_work} and our modeling of a CNN, we decompose a network cell with skip connections into a lattice network $\mathcal{G}$ and random network $\mathcal{R}$ (see Fig.~\ref{fig:skip_link}(c)).


\noindent\textbf{Proposed Metrics:} Our key objective is two-fold: (i) Quantify which topological characteristics of DNN architectures affect their performance, and (ii) Exploit such properties to accurately predict the test accuracy of a given architecture. To this end, we propose a new analytical metric called NN-Degree, as defined below.

\noindent\textbf{Definition of NN-Degree:} \textit{Given a DNN with $N_c$ cells, $d_c$ layers per cell, the width of each cell $w_c$, and the number of skip connections of each cell $SC_c$, the NN-Degree metric is defined as the sum of the average degree of each cell:}

\begin{equation}
\begin{split}
    g &=\sum_{c=1}^{N_c}(w_c +\frac{SC_c}{w_c\times d_c})\\ 
\end{split}
\label{eq:nn_deg}
\end{equation}

\noindent\textbf{Intuition:} The average degree of a given DNN cell is the sum of the average degrees from lattice network $\mathcal{G}$ and random network $\mathcal{R}$. Given a cell with $d_c$ convolutional layers and $w_c$ channels per layer, the number of nodes is $w_c\times d_c$. Moreover, each convolutional layer has $w_c\times w_c$ filters (kernels) accounting for the short-range connections; hence, in the lattice network $\mathcal{G}$, there are $w_c\times w_c\times d_c$ connections (total). Using the above analysis, we can express the NN-Degree as follows:
\begin{equation}
\begin{split}
    g &=g_\mathcal{G} + g_\mathcal{R}\\
    &=\sum_{c=1}^{N_c}\frac{number\ of\ connections\ in\ \mathcal{G}}{number\ of\ nodes\ in\ cell\ c} +\sum_{c=1}^{N_c} \frac{number\ of\ connections\ in\ \mathcal{R}}{number\ of\ nodes\ in\ cell\ c}\\
    &=\sum_{c=1}^{N_c}\frac{w_c\times d_c \times w_c}{w_c\times d_c} +\sum_{c=1}^{N_c} \frac{number\ of\ skip\ connections}{w_c\times d_c}\\
    &=\sum_{c=1}^{N_c}(w_c +\frac{SC_c}{w_c\times d_c})\\    
\end{split}
\label{eq:nn_deg_dev}
\end{equation}

\noindent\textbf{Discussion:} The first term in Equation~\ref{eq:nn_deg_dev} (i.e., $g_\mathcal{G}$) reflects the the width of the network $w_c$. Many successful DNN architectures, such as DenseNets~\cite{densenet}, Wide-ResNets~\cite{wide_resnet}, and MobileNets~\cite{mobilenetv2}, have shown that wider networks can achieve a higher test performance. The second term (i.e., $g_\mathcal{R}$) quantifies how densely the nodes are connected through the skip connections. As discussed in \cite{ensemble_resnet}, networks with more skip connections have more forward/backward propagation paths, thus have a better test performance. Based on the above analysis, we claim that a higher NN-Degree value should indicate networks with higher test performance. We verify this claim empirically in the experimental section. Next, we propose an accuracy predictor based only on the NN-Degree.

\vspace{2mm}
\noindent\textbf{Accuracy Predictor:} Given the NN-Degree ($g$) definition, we build the accuracy predictor by using a variant of logistic regression. Specifically, the test accuracy $\theta$ of a given architecture is: 
\begin{equation}
    \theta= \frac{1}{a_\theta+\text{exp}(b_\theta \times \frac{1}{g}+c_\theta)}\\
    \label{acc_predictor}
\end{equation}
where $a_\theta,b_\theta,c_\theta$ are the parameters that are fine-tuned with the accuracy and NN-Degree of sample networks from the search space. 
Section \ref{sec:experimental_results} shows that by using as few as 25 data samples (NN-Degree and corresponding accuracy values), we can generate an accurate predictor for a huge search space covering more than 63 billion configurations within 1 second on {a 20-core Intel Xeon CPU}.

\noindent\textbf{Training-free NAS:} Section~\ref{sec:experimental_results} shows that NN-Degree can indicate the test accuracy of a given architecture. Hence, one can use NN-Degree as a proxy of the test accuracy to enable the training-free NAS. Section~\ref{subsec:tf_nas} demonstrates that we can do \textit{training-free} NAS within 0.11 seconds on a 20-core CPU.

\subsection{Overview of In-memory Computing (IMC)-based Hardware}

\begin{figure}[t]
	\centering
	\includegraphics[width=0.88\textwidth]{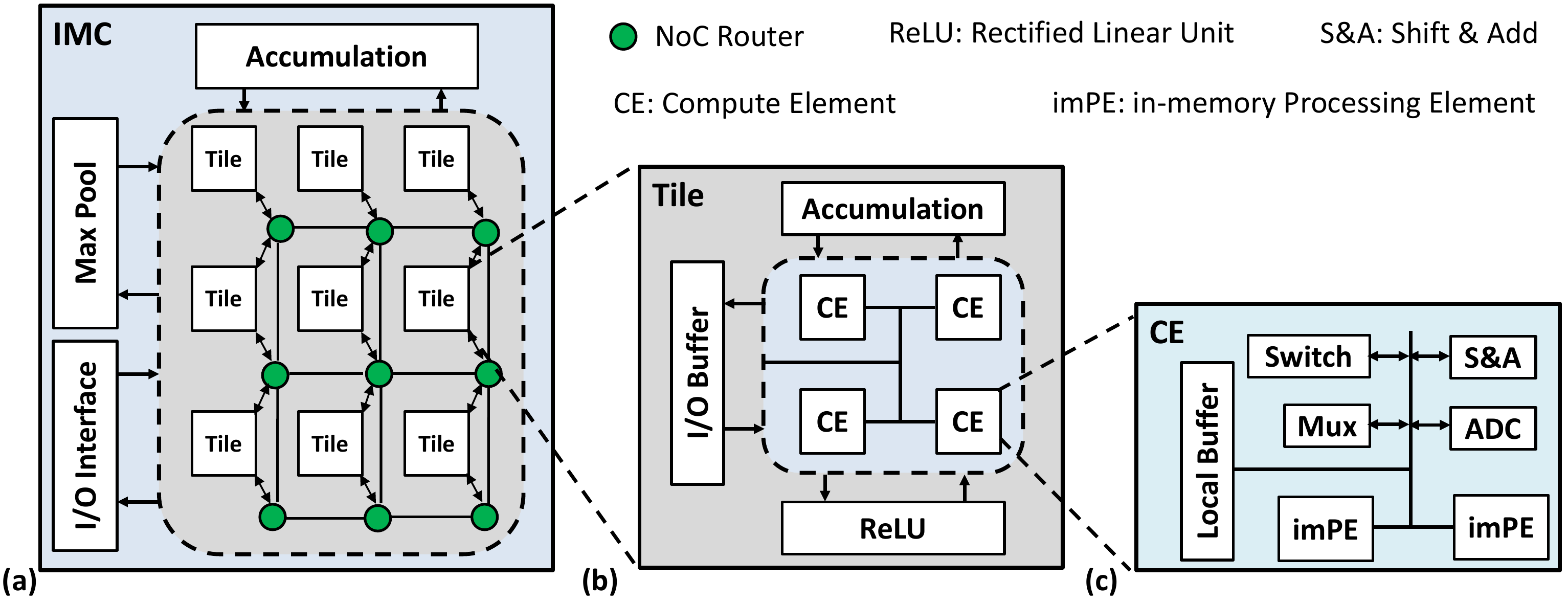}
    \caption{Details of the IMC hardware. (a) The architecture consists of multiple tiles connected via routers; (b) The structure of a tile. Each tile consists of multiple computing elements (CE), I/O buffer, ReLU unit and accumulation unit; (c) The structure of each CE. Each CE consists of multiple in-memory processing elements (imPE), local buffers, switch, multiplexer, analog to digital converter (ADC), shift and add (S\&A) circuit.}
	\label{fig:imc_arch}
\end{figure}

Fig.~\ref{fig:imc_arch} shows the IMC architecture considered in this work. We note that the proposed FLASH methodology is not specific to IMC-based hardware. We adopt an IMC architecture since it has been proven to achieve less memory access latency~\cite{horowitz20141}. Due to the high communication volume imposed by deeper and denser networks, communication between multiple tiles is crucial for hardware performance, as shown in~\cite{krishnan2020interconnect, mandal2020latency}.

Our architecture consists of multiple tiles connected by network-on-chip (NoC) routers, as shown in Fig.~\ref{fig:imc_arch}(a). We use a mesh-based  NoC due to its superior performance compared to bus-based architectures. Each tile consists of a fixed number of compute elements (CE), a rectified linear unit (ReLU), an  I/O buffer, and an accumulation unit, as shown in Figure Fig.~\ref{fig:imc_arch}(b).

Within each CE, there exist a fixed number of im-memory processing elements (imPE), a multiplexer, a switch, an analog-to-digital converter (ADC), a shift and add (S\&A) circuit, and a local buffer~\cite{chen2018neurosim}, as shown in Fig.~\ref{fig:imc_arch}(c). The ADC precision is set to four bits to avoid any accuracy degradation. There is no digital-to-analog (DAC) converter used in the architecture. A sequential signaling technique to represent multi-bit inputs is adopted~\cite{peng2019inference}. Each imPE consists of 256$\times$256 IMC crossbars (the memory elements) based on ReRAM (1T1R) technology ~\cite{krishnan2020interconnect, mandal2020latency, chen2018neurosim}.
This work incorporates a sequential operation between DNN layers since a pipelined operation may cause pipeline bubbles during inference~\cite{song2017pipelayer, qiao2018atomlayer}.

\begin{table}[htb]
\caption{Symbols and their corresponding definition used in our analytical area, latency, and energy models.}
\scalebox{0.86}{\begin{tabular}{|l|l||l|l|}
\hline
Symbol                                                                       & Definition                                                                                  & Symbol                                                           & Definition                                                                                             \\ \hline
\hline
$N_c$                                                                        & Number of cells                                                                             & $N_i^r$                                                          & \begin{tabular}[c]{@{}l@{}}Number of rows of imPE arrays\\of $i^\mathrm{th}$ layer\end{tabular}              \\ \hline
$a_\theta,b_\theta,c_\theta$                                                 & \begin{tabular}[c]{@{}l@{}}Learnable parameters for\\ accuracy predictor\end{tabular}       & $N_i^c$                                                          & \begin{tabular}[c]{@{}l@{}}Number of columns of imPE arrays\\ of $i^\mathrm{th}$ layer\end{tabular}           \\ \hline
$w_m$                                                                        & Width multiplier                                                                            & $Kx_i, Ky_i$                                                     & \begin{tabular}[c]{@{}l@{}}Kernel size \\of $i^\mathrm{th}$ layer\end{tabular}                         \\ \hline
$d_c$                                                                        & Number of layers within cell $c$                                                            & \begin{tabular}[c]{@{}l@{}}$N_i^{if}$,\ $N_i^{of}$\end{tabular} & \begin{tabular}[c]{@{}l@{}}Number of input and\\output features of  $i^\mathrm{th}$ layer\end{tabular} \\ \hline
$w_c$                                                                        & Width of cell $c$                                                                           & \begin{tabular}[c]{@{}l@{}}$(PE_x)_i$,\ $(PE_y)_i$\end{tabular} & \begin{tabular}[c]{@{}l@{}}Size of a single imPE \\ of $i^\mathrm{th}$ layer\end{tabular}                    \\ \hline
$SC_c$                                                                       & \begin{tabular}[c]{@{}l@{}}Number of skip connections\\ within cell $c$\end{tabular}        & $T_i$                                                            & \begin{tabular}[c]{@{}l@{}}Number of tiles\\of $i^\mathrm{th}$ layer\end{tabular}                     \\ \hline
$FLOP_c$                                                                     & Number of FLOPs of cell $c$                                                                 & $c$                                                              & \begin{tabular}[c]{@{}l@{}}Number of CEs in each tile\\ of $i^\mathrm{th}$ layer\end{tabular}          \\ \hline
$Comm_c$                                                                     & \begin{tabular}[c]{@{}l@{}}The amount of data transferred\\ through NoC inside cell $c$\end{tabular}              & $p$                                                              & Number of imPEs in each CE                                                                               \\ \hline
$N_T$                                                                        & \begin{tabular}[c]{@{}l@{}}Total number of tiles\\ of the chip\end{tabular}                 & $A_T$                                                            & Area of a tile                                                                                         \\ \hline
$F_{\mathcal{E}}$                                                            & Features for energy                                                                         & $\mathcal{E}^T$                                                  & \begin{tabular}[c]{@{}l@{}}Energy consumption\\ of a tile\end{tabular}                                 \\ \hline
\begin{tabular}[c]{@{}l@{}}$\Lambda_{comp}$, \ $\Lambda_{NoC}$\end{tabular} & \begin{tabular}[c]{@{}l@{}}Weight vectors to estimate\\ computation and NoC latency\end{tabular} & \begin{tabular}[c]{@{}l@{}}$F_{Comp}$,\ $F_{NoC}$\end{tabular}  & \begin{tabular}[c]{@{}l@{}}Features to estimate computation\\ and NoC latency\end{tabular}                     \\ \hline
\end{tabular}}
\label{table:hw_model}
\end{table}

\subsection{Hardware Performance Modeling}
This section describes the methodology of modeling hardware performance.
We consider three metrics for hardware performance: area, latency, and energy consumption. We use customized versions of NeuroSim~\cite{chen2018neurosim} for circuit simulation (computing fabric) and BookSim~\cite{jiang2013detailed} for cycle-accurate NoC simulation (communication fabric). First, we describe the details of the simulator.

\noindent\textbf{Input to the simulator:} The inputs to the simulator include the DNN structure, technology node, and frequency of operation. In this work, we consider a layer-by-layer operation. Specifically, we simulate each DNN layer and add its performance at the end to obtain the total performance of the hardware for the DNN.

\noindent\textbf{Simulation of computing fabric:} Table~\ref{tab:circuit_param} shows the parameters considered for the simulation of computing fabric. At the start of the simulation, the number of in-memory computing tiles is computed. Then, the area and energy of one tile are computed through analytical models derived from HSPICE simulation. After that, the area and energy of one tile are multiplied by the total number of tiles to obtain the total area and energy of the computing fabric. The latency of the computing fabric is computed as a function of the workload (the DNN being executed). We note that the original version of NeuroSim considers point-to-point on-chip interconnects, while our proposed work uses mesh-based NoC. Therefore, we skip the interconnect simulation in NeuroSim.

\noindent\textbf{Simulation of communication fabric:} We consider cycle-accurate simulation for the communic-ation fabric. BookSim is used to perform simulation. First, the number of tiles required for each layer is obtained from the simulation of computing fabric. In this work, we assume that each tile is connected to a dedicated router of the NoC. A trace file is generated corresponding to the particular layer of the DNN. The trace file consists of the information of the source router, destination router, and timestamp when the packet is generated. The trace file is simulated through BookSim to obtain the latency to finish all the transactions between two layers. We also obtain the area and energy of the interconnect through BookSim. Table~\ref{tab:circuit_param} shows the parameters considered for the interconnect simulator. More details of the simulator can be found in~\cite{krishnan2021interconnect}.

For hardware performance modeling, first we obtain the performance of the DNN through simulation, then the performance numbers are used to construct the performance models.

\noindent\textbf{Analytical Area Model:} An in-memory computing-based DNN accelerator consists of two major components: computation and communication.
The computation unit consists of multiple tiles and peripheral circuits; the communication unit includes an NoC with routers and other network components (e.g., buffers, links).
To estimate the total area, we first compute the number of rows ($N^r_i$) and number of columns ($N^c_i$) of imPEs required for the $i^\mathrm{th}$ layer of the DNN following Equation~\ref{eq:Nr} and Equation~\ref{eq:Nc}. 
%
\begin{equation}\label{eq:Nr}
    N^r_i = \Big\lceil \frac{Kx_i \times Ky_i \times N^{if}_i}{(PE_x)_i} \Big\rceil
\end{equation}
\begin{equation}\label{eq:Nc}
    N^c_i = \Big\lceil \frac{N^{of}_i \times N_{bits}}{(PE_y)_i} \Big\rceil
\end{equation}
where all the symbols are defined in Table \ref{table:hw_model}. Therefore, total number of imPEs required for the $i^\mathrm{th}$ layer of the DNN is $N^r_i \times N^c_i$.
Each tile consists of $c$ CEs, and each CE consists of $p$ number of imPEs.
Accordingly, each tile comprises $c \times p$ imPEs.
Therefore, the total number of tiles required for the $i^\mathrm{th}$ layer of the DNN ($T_i$) is:

\begin{equation}\label{eq:T}
    T_i = \Big\lceil \frac{N^r_i \times N^c_i}{c \times p} \Big\rceil
\end{equation}
Hence, the total number of tiles ($N_T$) required for a given DNN is $N_T=\sum_i T_i$.

\begin{table}[t]
\caption{Parameters used for simulation of computation and communication fabric.}
\scalebox{0.88}{\begin{tabular}{|l|l|l|l|}
\hline
\multicolumn{2}{|c|}{Circuit}           & \multicolumn{2}{c|}{NoC}      \\ \hline
imPE array size        & $128 \times 128$ & Bus width              & 32   \\ \hline
Cell levels          & 2 bit/cell       & Routing algorithm      & X--Y \\ \hline
Flash ADC resolution & 4 bits           & Number of router ports & 5    \\ \hline
Technology used      & RRAM            & Topology               & Mesh \\ \hline
\end{tabular}}
\label{tab:circuit_param}
\end{table}

As shown in Fig.~\ref{fig:imc_arch}(a), each tile is connected to the NoC routers for the on-chip communication. 
We assume that the total number of required routers is equal to the total number of tiles.
Hence, the total chip area is expressed as follows:

\begin{equation}
\begin{aligned}
    \mathcal{A} &= A_{comp} + A_{NoC} \\ 
    & = ( A_{Tile}^{Tot} + A_{Periphery} ) + ( A_{Router}^{Tot} + A_{others} ) \\ 
    & = N_T \times A_T + N_T \times A_R + ( A_{Periphery} + A_{others} )  \\ 
    & = N_T \times (A_T + A_R) + A_{rest} \\ \label{eq:area_model}
\end{aligned}
\end{equation}
where $A_{Tile}^{Tot}$ is the area accounted for all tiles and $A_{Router}^{Tot}$ is the total area accounted for all routers in the design. The area of a single tile is denoted by $A_{T}$; there are $N_T$ tiles in the design. Therefore $A_{Tile}^{Tot} = N_T \times A_T$.
The area of the peripheral circuit ($A_{Periphery}$) consists of I/O interface, max pool unit, accumulation unit, and global buffer.
The area of a single router is denoted by $A_{R}$; the number of routers is equal to the number of tiles ($N_T$). Therefore $A_{Router}^{Tot} = N_T \times A_R$.
The area of other components in the NoC ($A_{rest}$) comprises links and buffers.

\begin{figure}
    \centering
    \vspace{2mm}
    \includegraphics[width=0.88\textwidth]{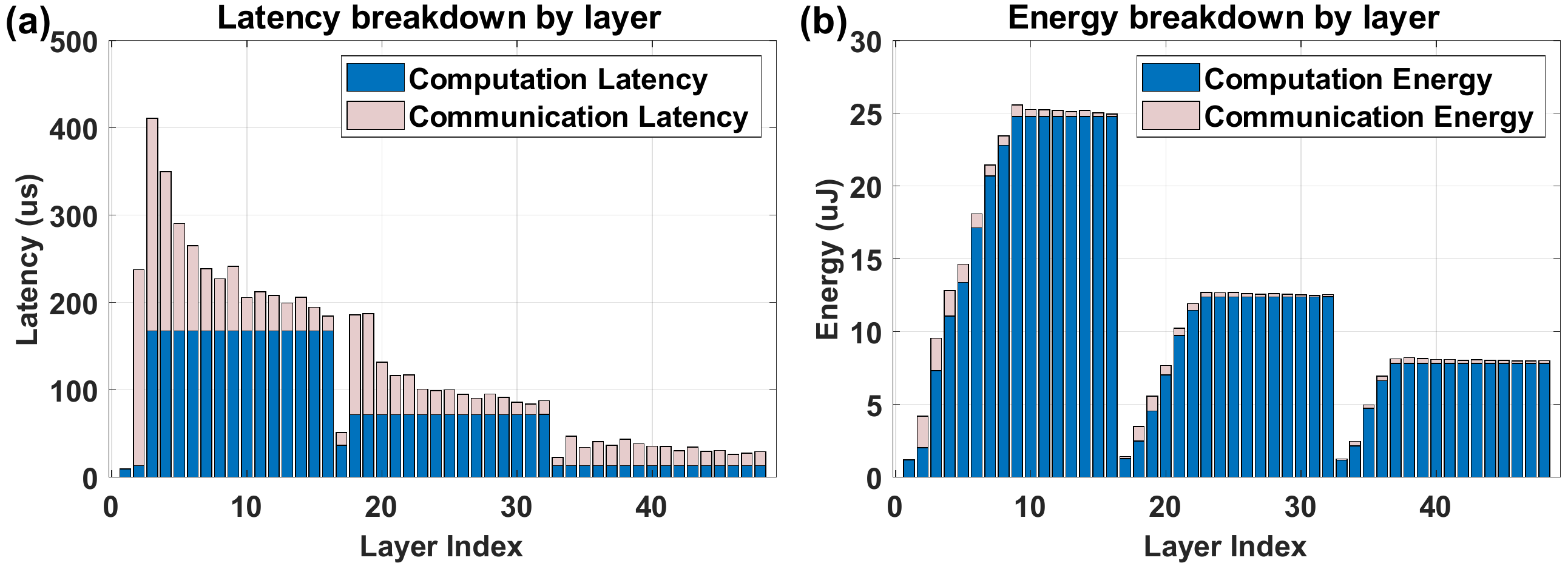}
    \caption{Layerwise hardware performance breakdown of a DNN  with 3 cells ($N_c=3$), 16 layers per cell ($d_c=16$), and a total of 48 layers. (a) Latency breakdown layer by layer: the computation latency accounts for 37.9\% of the total latency, while communication accounts for 62.1\%. (b) Energy consumption breakdown layer by layer: the computation energy accounts for 96.1\% of the total latency, while communication accounts for 3.9\%.}
    \label{fig:break}
\end{figure}

\noindent\textbf{Analytical Latency Model:} Similar to area, the total latency consists of computation latency and communication latency, as shown in Fig.~\ref{fig:break}(a).
To construct the analytical model of latency, we use floating-point operations (FLOPs) of the network to represent the computational workload.
We observe that the FLOPs of a given network are roughly proportional to the total number of convolution filters (kernels), which is the product of the number of layers and the square of the number of channels per layer (i.e., width value). 
In the network search space we consider, the width is equivalently represented by the width multiplier $w_m$ and the number of layers is $N_c\times d_c$; hence, we express the number of FLOPs of a given network approximately as the product of the number of layers, and the square of width multiplier: 
\begin{equation}
    FLOPs\sim  N_c d_c w_m^2
\end{equation}
Moreover, communication volume increases significantly due to the skip connections.
To quantify the communication volume due to skip connections, we define $Comm_c$ (the communication volume of a given network cell $c$) as follows:

$$Comm_c=SC_c \times Featuremap\ size\ of\ each\ SC$$
Combining the above analysis of computation latency and communication latency, we use a linear model to build our analytical latency model as follows:

\begin{equation}
    \mathcal{L}=\mathcal{L}_{comp} + \mathcal{L}_{NoC} = \Lambda^T_{comp} F_{comp} + \Lambda^T_{NoC} F_{NoC} 
\end{equation}
where $\Lambda^T_{comp}$ is a weight vector and $F_{comp} = [ w_m,d_c,N_c,N_cd_cw_m^2 ]$ is the vector of features with respect to the computation latency;
$\Lambda^T_{NoC}$ is another weight vector and $F_{comp} = [ SC_{c}, Comm_c ]$ is the vector of features corresponding to the NoC latency.
We randomly sample some networks from the search space and measure their latency to fine-tune the values of $\Lambda^T_{comp}$ and $\Lambda^T_{NoC}$.

\noindent\textbf{Analytical Energy Model:} We divide the total energy consumption into computation energy and communication energy, as shown in Fig.~\ref{fig:break}(b).
Specifically, the entire computation process inside each tile consists of three steps: 
\begin{itemize}
    \item Read the input feature map from the I/O buffer to the CE;
    \item Perform computations in CE and ReLU unit, then update the results in the accumulator;
    \item Write the output feature map to the I/O buffer.
\end{itemize}
Therefore, both the size of feature map and FLOPs contribute to the computation energy of a single cell.
Moreover, the communication energy consumption is primarily determined by the communication volume, i.e., ($Comm_c$). Hence, we use a linear combination of features to estimate the energy consumption of each tile $\mathcal{E}^T$:
%
%
\begin{equation}
    \mathcal{E}^T = \Lambda_\mathcal{E}^T F_\mathcal{E}
\end{equation}
where $\Lambda_\mathcal{E}^T$ is a weight vector and $F_\mathcal{E} = [w_m,d_c,N_c,SC_{c}, Comm_c, FLOP_c, FM_c]$ are the features corresponding to the energy consumption of each tile.
We use the measured energy consumption values of several sample networks to fine-tune the values of $\Lambda_\mathcal{E}^T$.
%
%
The total energy consumption ($\mathcal{E}$) is the product of $\mathcal{E}^T$ and number of tiles:
\begin{equation}
    \mathcal{E}=\Lambda_\mathcal{E}^T F_\mathcal{E} N_T
\end{equation}

%

We note that all the features used in both our accuracy predictor and analytical hardware performance model are only related to the architecture of the network through the basic parameters $\{w_m,d_c,N_c,SC_{c}\}$. 
Therefore, the analytical hardware models are lightweight. We note that there exist no other lightweight analytical models for IMC platforms.
Besides this, FLASH is general and can be applied to different hardware platforms.
For a given hardware platform, energy, latency, and area of the DNNs need to be first collected. Then the analytical hardware models need to be trained using the performance data.


\begin{algorithm}[t]
    \SetAlgoLined
    \KwInput{ \\
    \qquad Objective function: $f_{obj}$; \\ 
    \qquad Global search space: \\
    \qquad \quad $SP_{global}=[N_{cmin},N_{cmax}]\times [w_{m_{min}}, w_{m_{max}}] \times [d_{cmin}, d_{cmax}]\times [SC_{cmin}, SC_{cmax}]$; \\ 
    
    \qquad Search constraints: $S_{cons}=\{L_M, E_M, A_M,\theta_M\}$ ;\\
    \qquad Coarse-grain search step size: $\lambda $
    }
    
    \KwOutput{\\ \qquad The optimal architecture $\{w_m^{*},N_c^{*}, d_c^{*},SC_c^{*}\}$;\\ \\
    }
    
    \spprocess{ 
    Initialize Candidate Architecture Set ($CAS$) as empty set;

    \textbf{level 1: Fixed-$w_m$ Search}
    
    \For {$w_m$ \text{in} $[w_{m_{min}}, w_{m_{max}}]$}
    {   \textbf{level 2: Coarse-grain Search} 
    
        fix $w_m$, search the optimum $N_c^{G}, d_c^{G},SC_c^{G}$ with large search step $\lambda$ 
        
        $N_c^{G}, d_c^{G},SC_c^{G}$=SHGO(${f_{obj}}$, $SP_{global}$, $S_{cons}$, search step size=$\lambda$ ) 
        
            \qquad \textbf{level 3: Fine-grain Search} 
            
            \qquad within the neighbourhood of $N_c^{G}, d_c^{G},SC_c^{G}$, search the optimum $N_c^{L}, d_c^{L},SC_c^{L}$ 
            
            \qquad Local search space: $SP_{local}=\{ N_c^{G}\pm 2\lambda, d_c^{G}\pm 2\lambda,SC_c^{G}\pm 2\lambda\}$ 
            
            \qquad $N_c^{L}, d_c^{L},SC_c^{L}$=SHGO(${f_{obj}}$, $SP_{local}$, $S_{cons}$, search step size=$\mathrm{1}$ )

        Add $\{w_m,N_c^{L}, d_c^{L},SC_c^{L}\}$ to $CAS$
     }
     
    Compare the candidate architecture in $CAS$, find the optimum $\{w_m^{*},N_c^{*}, d_c^{*},SC_c^{*}\}$. 
    
    \textbf{Return} {$\{w_m^{*},N_c^{*}, d_c^{*},SC_c^{*}\}$} 
    
    }
    
    \caption{Our hierarchical SHGO-based search algorithm}
    \label{alg:shgo}
\end{algorithm}

\subsection{Optimal neural architecture search}

Based on the above accuracy predictor and analytical hardware performance models, we perform the second stage of our NAS methodology, i.e., searching for the optimal neural architecture by considering both test accuracy and hardware performance on the target hardware. To this end, we use a modified version of the Simplicial Homology Global Optimization (SHGO \cite{shgo}) algorithm to search for the optimum architecture. SHGO has mathematically rigorous convergence properties on non-linear objective functions and constraints and can solve derivative-free optimization problems\footnote{The detailed discussion of SHGO is beyond the scope of this paper. More details are available in \cite{shgo}}. Moreover, the convergence of SHGO requires much fewer samples and less time than reinforcement learning approaches \cite{jiang2020device}. 
Hence, we use SHGO for our new \textit{hierarchical} searching algorithm.

Specifically, as shown in Algorithm \ref{alg:shgo}, to further accelerate the searching process, we propose a \textit{three-level} SHGO-based algorithm instead of using the original SHGO algorithm.
At the first level, we enumerate $w_m$ in the search space. Usually, the range of $w_m$ is much more narrow than the other architecture parameters; hence without fixing $w_m$, we cannot use a large search step size for the second-level \textit{coarse-grain search}. 
At the second level, we use SHGO with a large search step size $\lambda$ to search for a coarse optimum $N_c^{G}, d_c^{G},SC_c^{G}$ 
by fixing the $w_m$. 
At the third level (\textit{fine-grain search}), we use SHGO with the smallest search step size (i.e., 1) to search for the optimum $N_c^{L}, d_c^{L},SC_c^{L}$ values for a specific $w_m$, within the neighborhood of the coarse optimum $N_c^{G}, d_c^{G},SC_c^{G}$, and add it to the candidate set. After completing the three-level search, we compare all neural architectures in the candidate set and determine the (final) optimal architecture $\{w_m^{*},N_c^{*}, d_c^{*},SC_c^{*}\}$.
To summarize, given the number of hyper-parameters $M$ and the number of possible values of each hyper-parameter $N$, the complexity of our hierarchical SHGO-based NAS is roughly proportional to MN, i.e., $O(MN)$.

Experimental results in Section \ref{sec:experimental_results} show that our proposed hierarchical search accelerates the overall search process without any decrease in the performance of the obtained neural architecture. Moreover, our proposed hierarchical SHGO-based algorithm involves much less computational workload compared to the original (one-level) SHGO-based algorithm and RL-based approaches~\cite{jiang2020device}; this even enables us to do NAS on a real Raspberry Pi-3B processor.

%% file: 4-experimental_results.tex
\section{Experimental Results}
\label{sec:experimental_results}
\subsection{Experimental setup}
\noindent\textbf{Dataset:} Existing NAS approaches show that the test accuracy of CNNs on CIFAR-10 dataset can indicate the test accuracy on other datasets, such as ImageNet~\cite{dong2020nasbench201}. Hence, similar to most of the NAS approaches, we use CIFAR-10 as the primary dataset. Moreover, we also evaluate our framework on CIFAR-100 and Tiny-ImageNet\footnote{Tiny-ImageNet is a downscaled-version ImageNet dataset with 64x64 resolution and 200 classes~\cite{img_net}. For more details, please check: \url{http://cs231n.stanford.edu/tiny-imagenet-200.zip}
} to demonstrate the generality of our proposed metric NN-Degree and accuracy predictor.

\noindent\textbf{Training Hyper-parameters:} We train each of the selected neural networks five times with PyTorch and use the mean test accuracy of these five runs as the final results. All networks are trained for 200 epochs with the SGD optimizer and a momentum of 0.9. We set the initial learning rate as 0.1 and use Cosine Annealing algorithm as the learning rate scheduler. 

\noindent\textbf{Search Space:} DenseNets are more efficient in terms of model size and computation workload than ResNets while achieving the same test accuracy~\cite{densenet}. Moreover, DenseNets have many more skip connections; this provides us with more flexibility for exploration compared to networks with Addition-type skip connections (ResNets, Wide-ResNets, and MobileNets). Hence, in our experiments, we explore the CNNs with DenseNet-type skip connections.

To enlarge the search space, we generate the generalized version of standard DenseNets by randomly selecting channels for concatenation. Specifically, for a given cell $c$, we define $t_c$ as the maximum skip connections that \textit{each layer} can have; thus, we use $t_c$ to control the topological properties of CNNs. Given the definition of $t_c$, layer $i$ can receive DenseNet-type skip connections (DTSC) from a maximum number of $t_c$ channels from previous layers within the same cell; that is, we randomly select $min\{w_c(i-1), t_c\}$ channels from layers ${0,1,...,(i-2)}$, and concatenate them at layer $i-1$. The concatenated channels then pass through a convolutional layer to generate the output of layer $i$ ($s_i$). Similar to recent NAS research~\cite{Darts}, we select links randomly because random architectures are often as competitive as the carefully designed ones. If the skip connections encompass all-to-all connections, this would result in the original DenseNet architecture~\cite{densenet}. An important advantage of the above setup is that we can control the number of DTSC (using $t_c$) to cover a vast search space with a large number of candidate DNNs. 

Like standard DenseNets, we can generalize this setup to contain multiple ($N_c$) cells of width $w_c$ and depth $d_c$; DTSC are present \textit{only within a cell} and not across cells. Furthermore, we increase the width (i.e., the number of output channels per layer) by a factor of 2 and halve the height and width of the feature map cell by cell, following the standard practice ~\cite{simonyan2014vgg}. After several cells (groups) of convolutions layers, the final feature map is average-pooled and passed through a fully-connected layer to generate the logits. The width of each cell is controlled using a width multiplier, $w_m$ (like in Wide-ResNets~\cite{wide_resnet}). The base number of channels of each cell is [16,32,64]. For $w_m = 3$, cells will have [48,96,192] channels per layer. To summarize, we control the value $\{w_m,N_c,d_c,t_{c}\}$ to sample candidate architectures from the entire search space.  
\begin{figure}[b]
    \centering
    \includegraphics[width=0.88\textwidth]{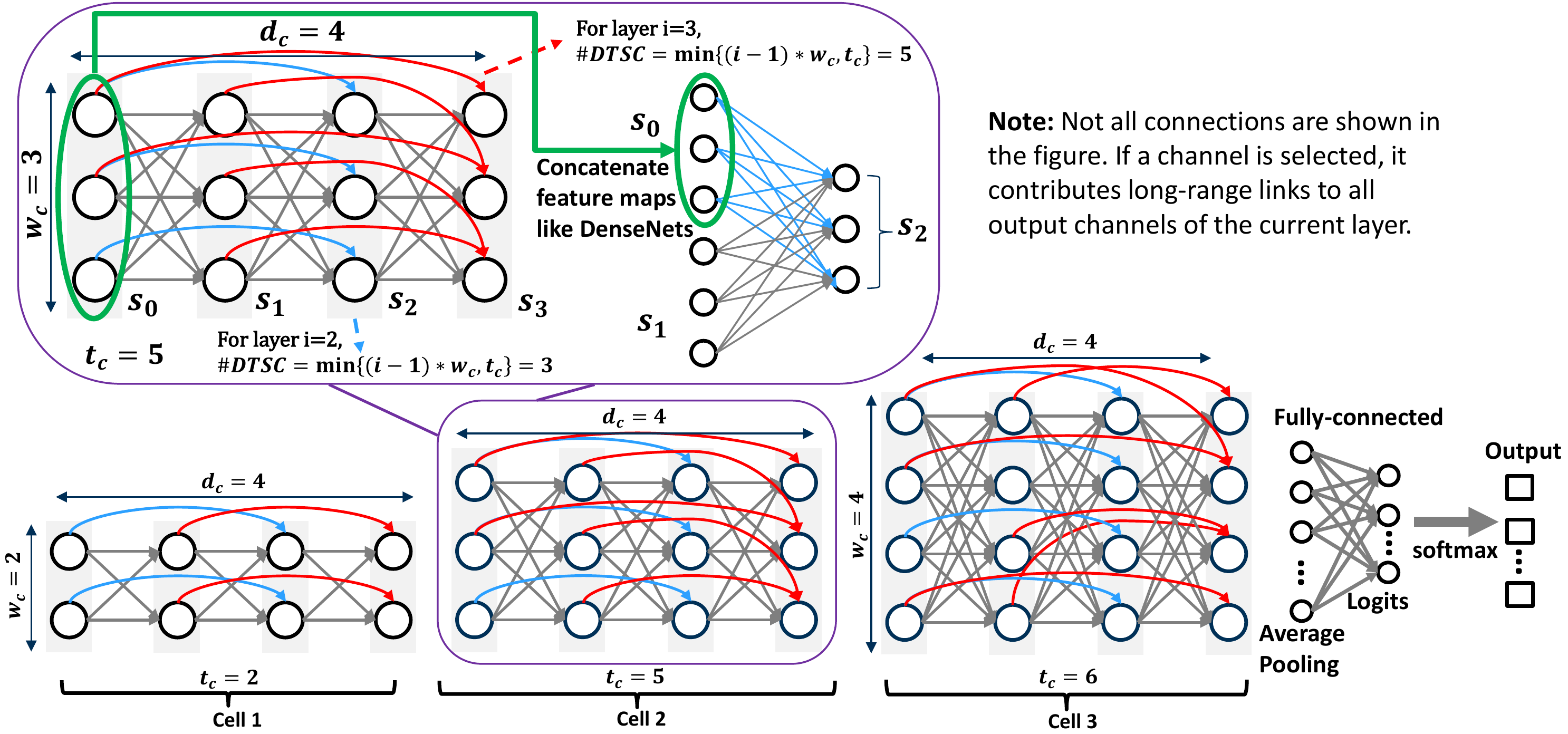}
    \vspace{-2mm}
    \caption{An example of candidate neural architectures from our search space. (The values of $w_c$ ,$t_c$, and $d_c$ are only for illustration and they do not represent the real search space). Not all skip connections are shown in the figure, for simplicity. The upper inset shows the contribution from all skip and short-range links to layer $i=2$: The feature maps for the randomly selected channels are concatenated as the input of the current layer $i=2$ (similar to DenseNets~\cite{densenet}). At each layer in a given cell, the maximum number of channels contributing to skip connections is controlled by $t_c$.}
    \label{fig:cnn_cell}
\end{figure}

Fig.~\ref{fig:cnn_cell} illustrates a sample CNN similar to the candidate architectures in our search space (small values of $w_c$ and $d_c$ are used for clarity). This CNN consists of three cells, each containing $d_c = 4$ convolutional layers. The three cells have a width (i.e., the number of channels per layer) of 2, 3, and 4, respectively. We denote the network width as $w_c = [2,3,4]$. Finally, the maximum number of channels that can supply skip connections is given by $t_c = [2,5,6]$. 
That is, the first cell can have a maximum of two skip connection candidates per layer (i.e., previous channels that can supply skip connections), the second cell can have a maximum of five skip connections candidates per layer, and so on. Moreover, as mentioned before, we randomly choose $min\{w_c(i-1), t_c\}$ channels for skip connections at each layer. The inset of Fig. \ref{fig:cnn_cell} shows for a specific layer, how skip connections are created by concatenating the feature maps from previous layers.

In practice, we use three cells for the CIFAR-10 dataset, i.e., $N_c=3$. We constrain the $1\leq w_m\leq 3$ and $5\leq d_c\leq 30$. We also constrain $t_c$ of each cell: $5\leq t_{1}$, $2t_{1}\leq t_{2}$ and $2t_{2}\leq t_{3}$ for these three cells, respectively. In this way, we can balance the number of skip connections across each cell. Moreover, the maximum number of skip connections that a layer can have is the product of the width of the cell ($w_c$) and $d_c-2$ which happens for the last layer in a cell concatenating all of the output channels except the second last layer. Hence, the upper bound of $t_c$, for each cell, is $16w_m(d_c -2),32w_m(d_c -2),64w_m(d_c -2)$, respectively. Therefore, the size of the overall search space is:
$$\sum_{w_m=1}^{3}\sum_{d_c =5}^{30}\sum_{t_{1} =5}^{16w_m(d_c -2)}\sum_{t_{2} =2t_{1}}^{32w_m(d_c -2)}({64w_m(d_c -2)- 2t_{2}}+1)=6.39 \times 10^{10}$$
\noindent\textbf{Hardware Platform:} The training of the sample neural architectures from the search space is conducting on Nvidia GTX-1080Ti GPU. We use Intel Xeon 6230, a 20-core CPU, to simulate the hardware performance of multiple candidate networks and fine-tune the accuracy predictor and analytical hardware models. Finally, we use the same 20-core CPU to conduct the NAS process.

\subsection{Accuracy Predictor}

\begin{figure} [b]
    \centering
    \includegraphics[width=0.96\textwidth]{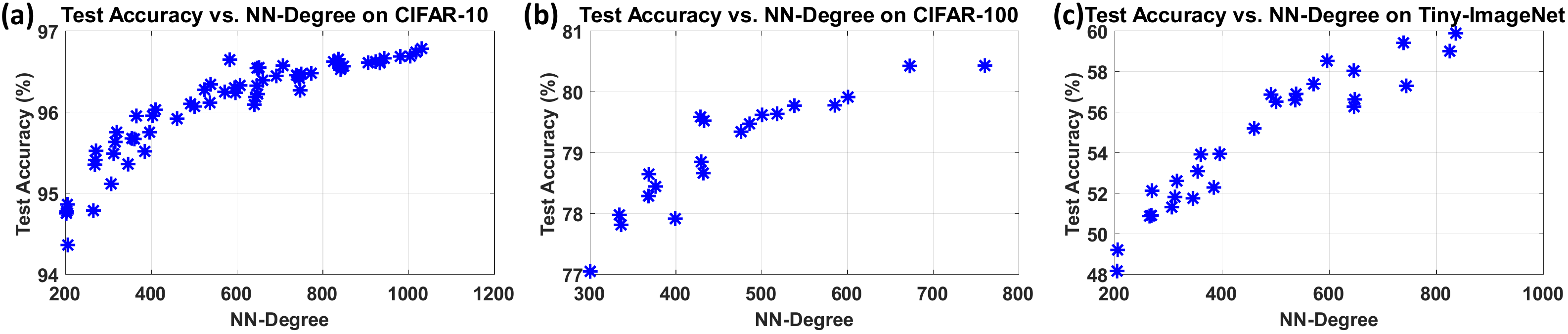}
    \vspace{-4mm}
    \caption{We randomly select multiple networks from the search space then train and test their accuracy on \textbf{CIFAR-10, CIFAR-100, Tiny-ImageNet} datasets. (a) Real test accuracy vs. NN-Degree: networks with higher NN-Degree values have a higher test accuracy on the \textbf{CIFAR-10} dataset. (b) Real test accuracy vs. NN-Degree: networks with higher NN-Degree values have a higher test accuracy on the \textbf{CIFAR-100} dataset. (c) Real test accuracy vs. NN-Degree: networks with higher NN-Degree values have a higher test accuracy on the \textbf{Tiny-ImageNet} dataset. }
    \label{fig:acc_deg}
\end{figure}

\begin{figure} [b]
    \centering
    \includegraphics[width=0.96\textwidth]{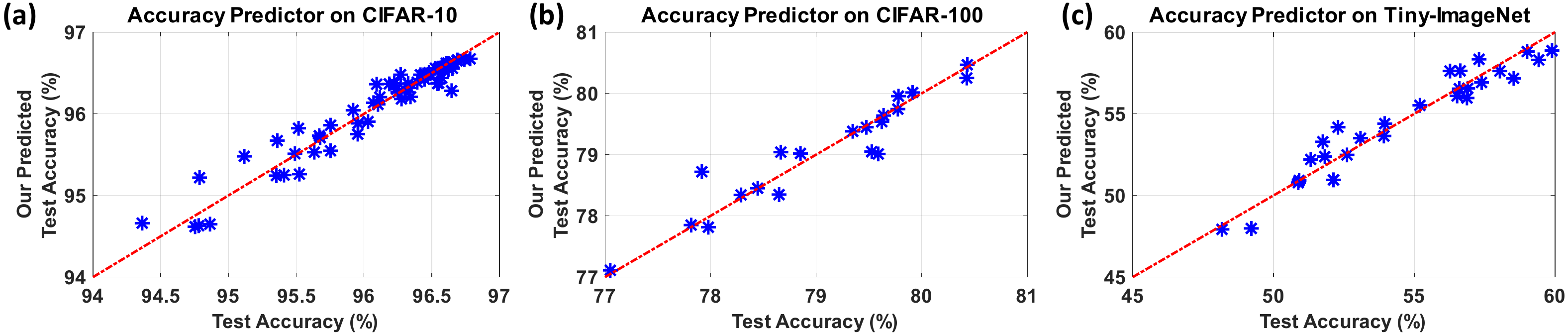}
    \vspace{-4mm}
    \caption{(a) Predictions of our NN-Degree based accuracy predictor vs. real test accuracy on \textbf{CIFAR-10} dataset. (b) Predictions of our NN-Degree based accuracy predictor vs. real test accuracy on \textbf{CIFAR-100} dataset. (c) Predictions of our NN-Degree based accuracy predictor vs. real test accuracy on \textbf{Tiny-ImageNet} dataset. The red dotted lines in these figures show a very good correlation between the predicted and measured values.}
    \label{fig:acc_predict}
\end{figure}

\begin{table}[t]
\caption{Our NN-Degree based accuracy predictor for neural architecture search vs. existing predictors implemented by graph-based neural networks. We calculate the improvement ratio for each of the metric by considering the best among all existing approaches in this table. (`-'  denotes that the corresponding results are not reported or not applicable.)}
\vspace{-3mm}
\footnotesize
\setlength\tabcolsep{3.5pt}
\begin{tabular}{|l|l|l|l|l|l|l|l|}
\hline
\multicolumn{1}{|c|}{\multirow{2}{*}{\begin{tabular}[c]{@{}c@{}}Accuracy Estimation\\ Technique\end{tabular}}} & \multicolumn{2}{c|}{Search Space (SS) Size}                                                                                     & \multicolumn{2}{c|}{\# Training Samples}                                                                                         & \multicolumn{1}{c|}{\multirow{2}{*}{RMSE (\%)}} & \multicolumn{2}{c|}{Training Time (s)}                                                  \\ \cline{2-5} \cline{7-8} 
\multicolumn{1}{|c|}{}                                                                                         & \multicolumn{1}{c|}{Value}  & \multicolumn{1}{c|}{\% of FLASH SS } & \multicolumn{1}{c|}{Value} & \multicolumn{1}{c|}{\begin{tabular}[c]{@{}c@{}}Ratio ($\times$)\\ w.r.t FLASH \end{tabular}} & \multicolumn{1}{c|}{}                           & \multicolumn{1}{c|}{Value} & \multicolumn{1}{c|}{\begin{tabular}[c]{@{}c@{}}Ratio ($\times$)\\ w.r.t FLASH\end{tabular}} \\ \hline
GNN+MLP~\cite{eccv_gates}                                                                                                       & $4.2 \times 10^5$           & $6.6 \times 10^{-4}$ \%                                                                                   & $3.8 \times 10^5$          & 15250                                                                                  & -                                               & -                          & -                                                                                                   \\ \hline
GNN~\cite{pr_2020_gnn_acc_pre}                                                                                                            & $4.2 \times 10^5$           & $6.6 \times 10^{-4}$ \%                                                                                   & $3.0 \times 10^5$          & 11862                                                                                  & 0.05                                            & -                          & -                                                                                                   \\ \hline
GCN~\cite{brp_nas}                                                                                                           & $1.6 \times 10^4$           & $2.5 \times 10^{-5}$ \%                                                                                   & $1.0 \times 10^3$          & 40                                                                                                & \textgreater{}1.8                               & -                          & -                                                                                                   \\ \hline
GCN~\cite{yiran_gnn}                                                                                                           & $4.2 \times 10^5$           & $6.6 \times 10^{-4}$ \%                                                                                   & $1.7 \times 10^2$          & 6.88                                                                                                & 1.4                                             & 25                         & 66                                                                                                  \\ \hline
\textbf{\begin{tabular}[c]{@{}l@{}}FLASH (NN-Degree +\\ Logistic Regression) \end{tabular}}                & $\mathbf{6.4 \times 10^{10}}$ & \textbf{100\%}                                                                                          & $\mathbf{2.5 \times 10^1}$ & \textbf{1}                                                                                          & \textbf{0.152}                                   & \textbf{0.38}              & \textbf{1}                                                                                          \\ \hline
\end{tabular}
\label{tab:acc_predict}
\end{table}

%
We first derive the NN-Degree ($g$) for the neural architecture in our search space. Based on Equation \ref{eq:nn_deg}, we substitute $SC_c$ with the real number of skip connections in a cell as follows:
\begin{equation}
\begin{split}
    g =\sum_{c=1}^{N_c}(w_c +\frac{SC_c}{w_c\times d_c})   
    = \sum_{c=1}^{N_c} (w_c +\frac{\sum_{i=2}^{d_c-1} \text{min}\{(i-1)w_c,t_c\}}{d_c} ) 
\end{split}
\end{equation}
In Section \ref{sec:methodology}, we argue that the neural architecture with a higher NN-degree value tends to provide a higher test accuracy. In Fig. \ref{fig:acc_deg}(a), we plot the test accuracy vs. NN-Degree of 60 randomly sampled neural networks from the search space for CIFAR-10 dataset; our proposed network-topology based metric NN-Degree indicates the test accuracy of neural networks. Furthermore, Fig~\ref{fig:acc_deg}(b) and Fig~\ref{fig:acc_deg}(c) also show the test accuracy vs. NN-Degree of 20 networks on CIFAR-100 dataset and 27 networks on Tiny-ImageNet randomly sampled from the search space. Clearly, our proposed metric NN-Degree predicts the test accuracy of neural networks on these two datasets as well. Indeed, the results prove that our claim in Section \ref{sec:methodology} is empirically correct, i.e., networks with higher NN-Degree values have a better test accuracy. 

Next, we use our proposed NN-Degree to build the analytical accuracy predictor. We train as few as 25 sample architectures randomly sampled from the entire search space and record their test accuracy and NN-Degree on CIFAR-10, CIFAR-100, and Tiny-ImageNet datasets. Then, we fine-tune our NN-Degree based accuracy predictor described by Equation \ref{fig:acc_predict}. As shown in Fig. \ref{fig:acc_predict}(a), Fig~\ref{fig:acc_predict}(b), and Fig~\ref{fig:acc_predict}(c), our accuracy predictor achieves very high performance while using surprisingly few samples with only three parameters on all these datasets. 

We also compare our NN-Degree-based accuracy predictor with the current state-of-the-art approaches. As shown in Table~\ref{tab:acc_predict}, most of the existing approaches use Graph-based neural networks to make predictions~\cite{yiran_gnn,pr_2020_gnn_acc_pre,brp_nas,eccv_gates}. However, Graph-based neural networks require much more training data, and they are much more complicated in terms of computation and model structure compared to classical methods like logistic regression. Due to the significant reduction in the model complexity, our predictor requires $6.88\times$ fewer training samples, although a much larger search space ($1.5\times 10^5$ larger than the existing work) is covered. Moreover, our NN-Degree based predictor has only three parameters to be updated; hence it consumes $66\times$ less fine-tuning time than the existing approaches. Finally, besides such low model complexity and fast training process, our predictor achieves a very small RMSE (0.152\%) as well. 

During the search of our NAS methodology, we use the accuracy predictor to directly predict the accuracy of sample architectures as opposed to performing the time-consuming training. The high precision and low complexity of our proposed accuracy predictor also enable us to adopt very fast optimization methods during the search stage. 
Furthermore, because our proposed metric NN-Degree can predict the test performance of a given architecture, we can use NN-Degree as the proxy of the test accuracy to do NAS without the time-consuming training process. This \textit{training-free} property allows us to quickly compare the accuracy of given architectures and thus accelerate the entire NAS.

\begin{figure}[t]
	\centering
	\includegraphics[width=0.96\textwidth]{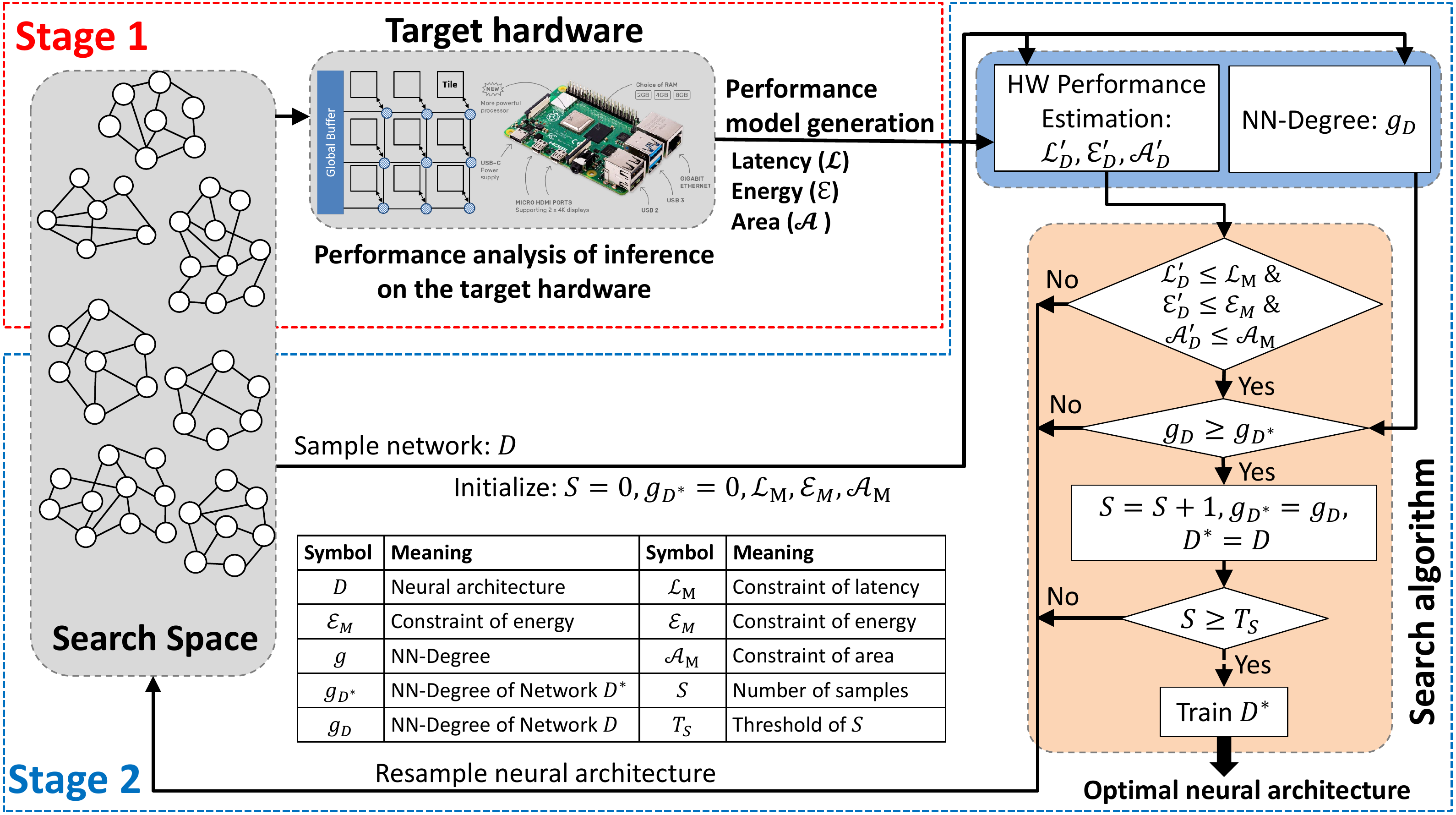}
	\vspace{-3mm}
    \caption{Overview of the proposed training-free NAS approach. Stage 1 (red box): we build hardware (HW) performance models by randomly sampling candidate networks from the search space to evaluate the hardware characteristics (latency $\mathcal{L}$, energy $\mathcal{E}$, and area $\mathcal{A}$). Stage 2 (blue box): we search for the optimal network architecture with the hardware performance constraints (i.e., $\mathcal{L}_M$, $\mathcal{E}_M$, and $\mathcal{A}_M$); we randomly choose some architectures and use the HW performance models to estimate their hardware performance. Then, we select the neural architecture $D^*$ with the highest NN-Degree which meets the HW performance constraints. Finally, we train the obtained architecture $D^*$ to get the optimal neural architecture. 
}
	\label{train_free_nas}
\end{figure}
\subsection{NN-Degree based Training-free NAS}\label{subsec:tf_nas}
\begin{table}

\caption{Our NN-Degree based training-free NAS (\textbf{FLASH}) and several representative time-efficient NAS on CIFAR-10 Dataset. We select the optimal architectures with the highest NN-Degree values among 20,000 randomly sampled architectures on a 20-core CPU.}
\vspace{-3mm}
\scalebox{0.88}{
{
\begin{tabular}{|l|l|l|l|l|l|}

\hline

Method& Search Method& {\#Params} & {Search Cost} & {Training needed} & {Test error (\%)} \\
\hline\hline
ENAS\cite{hyper_nas}& RL+weight sharing & 4.6M & 12 GPU hours& Yes & 2.89\\\hline
SNAS\cite{xie2018snas}& gradient-based & 2.8M & 36 GPU hours& Yes & 2.85\\\hline
DARTS-v1\cite{Darts} & gradient-based & {3.3M}& {1.5 GPU hours}& {Yes}& {3.0}\\\hline
DARTS-v2\cite{Darts} & gradient-based & {3.3M}& {4 GPU hours}& {Yes}& {2.76} \\\hline
ProxylessNAS\cite{cai2018proxylessnas} & gradient-based & {5.7M}& {NA} & {Yes}& {2.08} \\\hline
Zero-Cost\cite{tf_nas1} & Proxy-based& {NA}& {NA} & {Yes}& {5.78} \\\hline
TE-NAS\cite{tf_nas2} & Proxy-based & {3.8M}& {1.2 GPU hours} & {No}& {2.63} \\\hline

{\textbf{FLASH}}& \textbf{NN-Degree based}& {\textbf{3.8M}} & {\textbf{0.11 seconds}} & {\textbf{No}}& {\textbf{3.13}} \\
\hline

\end{tabular}}}
\label{notrain_tab}
\end{table}
To conduct the training-free NAS, we reformulate the problem described by Equation~\ref{eq:problem_definition} as follows:
\begin{equation}
        \max \theta,\quad \text{subject\ to:} \ \mathcal{A} \leq \mathcal{A}_M,
     \ \mathcal{L} \leq \mathcal{L}_M,
     \ \mathcal{E} \leq \mathcal{E}_M\\
\label{eq:problem_definition_raw}
\end{equation}

\noindent{To maximize the values of $\theta$, we can search for the network with maximal \textit{NN-Degree} values, which eliminate the training time of candidate architectures. In Fig.~\ref{train_free_nas}, we show how we can use the NN-Degree to do training-free NAS. During the first stage, we profile a few networks on the target hardware and fine-tune our hardware performance models. During the second stage, we randomly sample candidate architectures and select those which meet the hardware performance constraints. We use the fine-tuned analytical models to estimate the hardware performance instead of doing real inference, which improves the time efficiency of the entire NAS. After that, we select the optimal architecture with the highest NN-Degree values which meets the hardware performance constraints. We note that the NAS process itself is training-free (hence lightweight), as only the final solution $D^*$ needs to be trained.
}

To evaluate the performance of our training-free NAS framework, we randomly sample 20,000 candidate architectures from the search space and select the one with the highest NN-Degree values as the obtained/optimal architecture. Specifically, it takes only 0.11 seconds to evaluate these 20,000 samples’ NN-Degree on a 20-core CPU to get the optimal architecture (no GPU needed). As shown in Table~\ref{notrain_tab}, the optimal architecture among these 20,000 samples achieves a comparable test performance with the representative time-efficient NAS approaches but with much less time cost and computation capacity requirement.

\begin{figure} [b]
    \centering
    \includegraphics[width=1\textwidth]{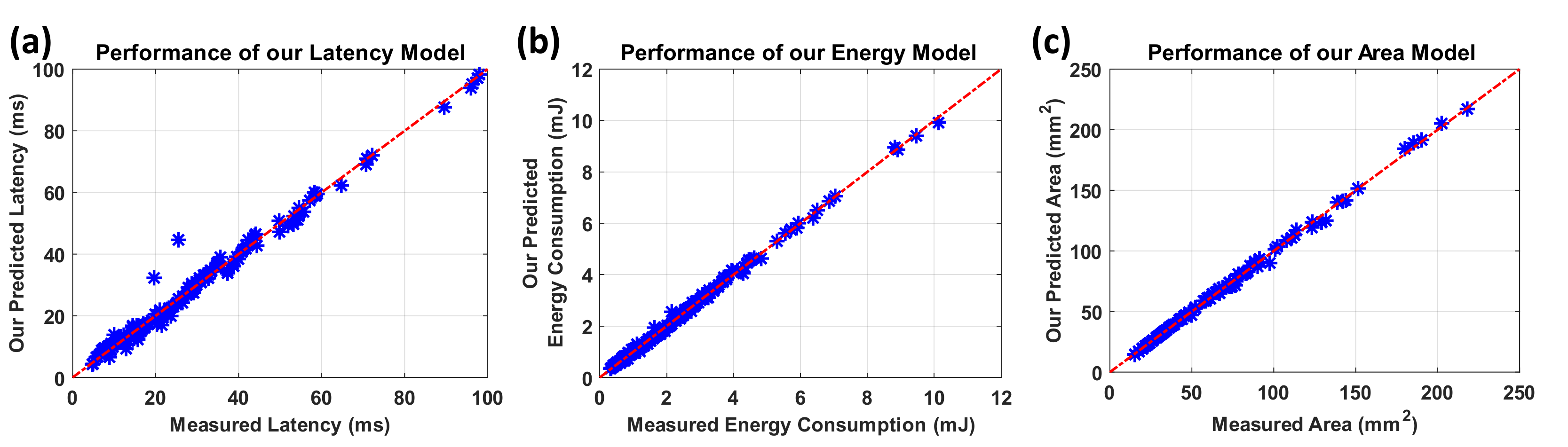}
    \vspace{-8mm}
    \caption{Performance of our analytical hardware models on \textbf{ImageNet} classification networks: (a) Predicted values by our analytical area model vs. measured area. (b) Predicted values by our analytical latency model vs. measured latency. (c) Predicted values by our analytical energy model vs. measured energy consumption. The red lines demonstrate that our proposed models generalize well for networks evaluated on ImageNet-scale datasets.}
    \label{fig:img_hw_model_perf}
\end{figure}

\subsection{Analytical hardware performance models}
Our experiments show that using 180 samples offers a good balance between the analytical models' accuracy and the number of fine-tuning samples. Hence, we randomly select 180 neural architectures from the search space to build our analytical hardware performance models. Next, we perform the inference of these selected 180 networks on our simulator~\cite{krishnan2021interconnect} to obtain their area, latency, and energy consumption. After obtaining the hardware performance of 180 sample networks, we fine-tune the parameters of our proposed analytical area, latency, and energy models discussed in Section \ref{sec:methodology}. To evaluate the performance of these fine-tuned models, we randomly select another 540 sample architectures from the search space then conduct inference and obtain their hardware performance.


Table~\ref{tab:hw_model_perf} summarizes the performance of our analytical models. The mean estimation error is always less than 4\%. Fig. \ref{fig:img_hw_model_perf} shows the estimated hardware performance obtained by our analytical model for the ImageNet dataset. We observe that the estimation coincides with the measured values from simulation.
Our analytical models enable us to obtain very accurate predictions of hardware performance with the time cost of less than 1 second on a 20-core CPU. The high performance and low computation workload enable us to directly adopt these analytical models to accelerate our searching stage instead of conducting real inference.

\begin{table}
\caption{Summary of the performance of our proposed analytical models for Area, Latency, and Energy. }
\scalebox{0.76}{

\begin{tabular}{|l|l|l|l|l|}
\hline
 Model& \#Features & Mean Error (\%) & Max Error (\%)  & Fine-tuning Time (s)\\ \hline\hline
 Area& 2 & 0.1 & 0.2 & 0.49 \\ \hline
 Latency& 9 & 3.0 & 20.8 & 0.52 \\ \hline
 Energy& 16 &3.7 & 24.4 &  0.56 \\ \hline
\end{tabular}}
\label{tab:hw_model_perf}
\end{table}

\begin{table}[t]
\caption{Estimation error with different ML models for \textbf{ImageNet} with IMC as target hardware platform.}
\scalebox{0.76}{\begin{tabular}{|l|l|l|l|}
\hline
                       & SVM   & \begin{tabular}[c]{@{}c@{}}Random Forest\\ (Max. Depth = 16)\end{tabular} & \begin{tabular}[c]{@{}c@{}}Analytical Models\\ (Proposed)\end{tabular} \\ \hline
Latency Est. Error (\%) & 58.98 & 8.23                                                                      & 6.7                                                                    \\ \hline
Energy Est. Error (\%)  & 78.49 & 11.01                                                                     & 3.5                                                                    \\ \hline
Area Est. Error (\%)    & 36.99 & 13.37                                                                     & 1.7                                                                    \\ \hline
\end{tabular}}
\label{tab:other_ml}
\end{table}

\begin{figure} [b]
    \centering
    \vspace{3mm}
    \includegraphics[width=1\textwidth]{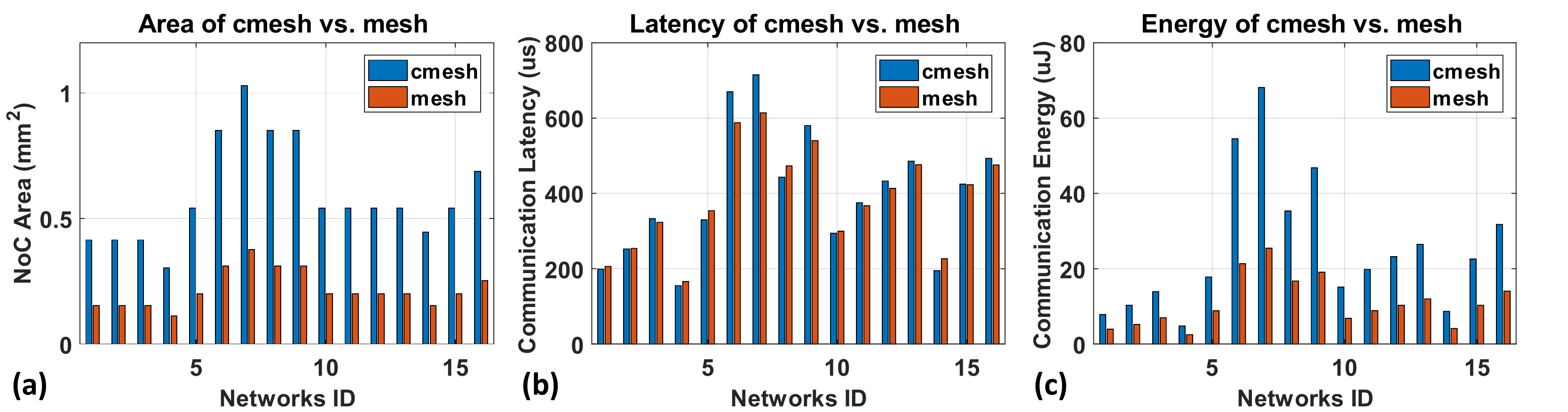}
    \caption{Performance comparison between our mesh-NoC and cmesh-NoC~\cite{shafiee2016isaac} on \textbf{CIFAR-10} classification networks for 16 different networks: (a) Our mesh-NoC needs much less area than the cmesh-NoC; (b) Our mesh-NoC has almost the same latency as the cmesh-NoC; (c) Our mesh-NoC consumes much less energy consumption than the cmesh-NoC.}
    \label{fig:cmesh_vs_mesh}
\end{figure}

\begin{figure} [htb]
    \centering
    \vspace{2mm}
    \includegraphics[width=1\textwidth]{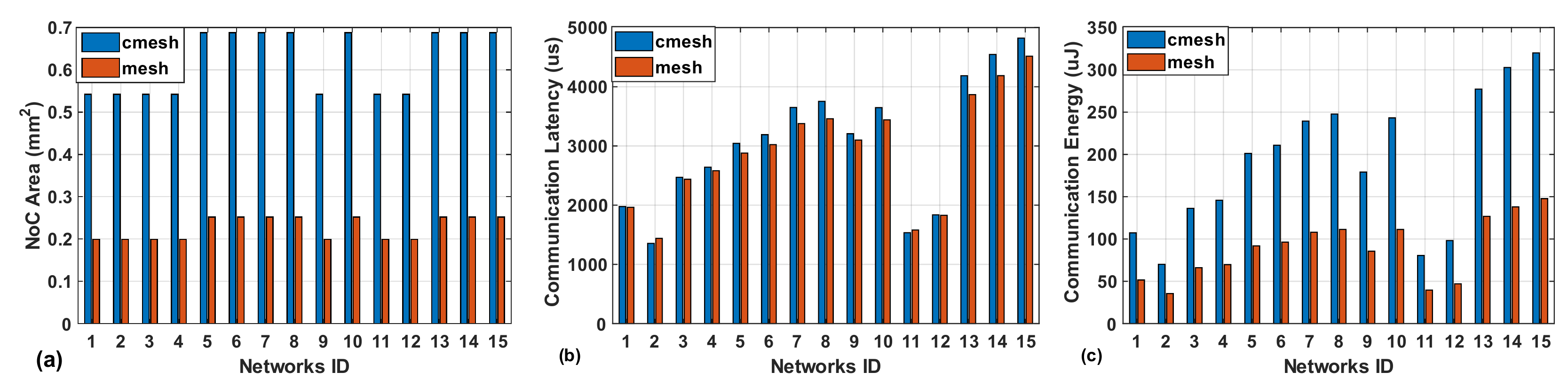}
    \vspace{-6mm}
    \caption{Performance comparison between our mesh-NoC and cmesh-NoC~\cite{shafiee2016isaac} on \textbf{ImageNet} classification networks for 15 different networks: (a) Our mesh-NoC needs much less area than the cmesh-NoC; (b) Our mesh-NoC has almost the same latency as the cmesh-NoC; (c) Our mesh-NoC consumes much less energy consumption than the cmesh-NoC.}
    \label{fig:cmesh_vs_mesh_img}
\end{figure}

\noindent\textbf{Comparison with other machine learning models:} Table~\ref{tab:other_ml} compares the estimation error for SVM, random forest with a maximum tree depth of 16 and the proposed analytical hardware models for ImageNet dataset. A maximum tree depth of 16 is chosen for random forest since it provides the best accuracy among random forest models. We observe that our proposed analytical hardware models achieve the smallest error among all three modeling techniques. SVM performs poorly since it tries to classify the data with a hyper-plane, and no such plane may exist given the complex relationship between the features and performance of the hardware platform.

\subsection{On-chip communication optimization}

As shown in Fig.~\ref{fig:cmesh_vs_mesh} and Fig.~\ref{fig:cmesh_vs_mesh_img}, we compare the NoC performance (area, energy, and latency) of our FLASH with respect to the cmesh-NoC~\cite{shafiee2016isaac} for 16 randomly selected networks from the search space for CIFAR-10 dataset and ImageNet dataset, respectively.
We observe that the mesh-NoC occupies on average only 37\% area and consumes only 41\% energy with respect to the cmesh-NoC.
Since the cmesh-NoC uses extra links and repeaters to connect diagonal routers, the area and energy with the cmesh-NoC are significantly higher than the mesh-NoC.
Additional links and routers in the cmesh-NoC result in lower hop counts than the mesh-NoC. However, the lower hop count reduces the latency at low congestion.
As the congestion in the NoC increases, the latency of the cmesh-NoC becomes higher than the mesh-NoC due to increased utilization of additional links. This phenomenon is also demonstrated in~\cite{grot2008scalable}.
Therefore, the communication latency with the cmesh-NoC is higher than the mesh-NoC for most of the DNNs.
The communication latency with the mesh-NoC is on average within 3\% different from the communication latency with the cmesh-NoC.
Moreover, we observe that the average utilization of the queues in the mesh-NoC varies between 20\%-40\% for the ImageNet dataset.
Furthermore, the maximum utilization of the queues ranges from 60\% to 80\%.
Therefore, the mesh-NoC is heavily congested.
Thus, our proposed communication optimization strategy outperforms the state-of-the-art approaches.

\subsection{Hierarchical SHGO-based neural architecture search}\label{sec:res_nas}
After we fine-tune the NN-Degree based accuracy predictor and analytical hardware performance models, we use our proposed hierarchical SHGO-based searching algorithm to do the neural architecture search. 

\noindent\textbf{Baseline approach:} Reinforcement Learning (RL) is widely used in NAS~\cite{jiang2020device, hsu2018monas, cell_1}; hence we have implemented a RL-based NAS framework as a baseline.
For the baseline, we consider the objective function in Equation~\ref{eq:problem_definition}.
Specifically, we incorporate a deep-Q network approach for the baseline-RL~\cite{mnih2013playing}.
We construct four different controllers for the number of cell ($N_c$), cell depth ($d_c$), width multiplier ($w_m$) and number of long skip connections ($SC_c$).
The training hyper-parameters for the baseline-RL are shown in Table~\ref{tab:params_RL}.
The baseline-RL approach estimates the optimal parameters ($N_c, d_c, w_m, SC_c$).
We tune the baseline-RL approach to obtain the best possible results. We also implement a one-level SHGO algorithm (i.e., original SHGO) as another baseline to show the efficiency of our hierarchical algorithm. 

\begin{table}[t]
\caption{Parameters chosen for the baseline-RL approach.}
\scalebox{0.76}{
\begin{tabular}{|l|l||l|l|}
\hline
Metric                & Value   & Metric                & Value\\ \hline
\hline
Number of layers      & 3      & Learning rate \ \ \ \ \ \ \ \ \ \ \ \ \ \ \ \ \ \ \ \ \ \ \ \ \ \ \ \ \ \ \ \ \ \ \ \       & 0.001   \\ \hline
Number of neurons in each layer & 20    &   Activation            & softmax \\ \hline
Optimizer             & ADAM   &  Loss            & MSE \\ \hline

\end{tabular}}
\label{tab:params_RL}
\end{table}

\begin{table} [b]
\caption{Comparison between RL-based search, one-level SHGO-based search, and our proposed hierarchical SHGO-based search. No constraint means that we don't set any bounds for the accuracy, area, latency, and energy consumption of the networks; we compare FLASH with RL when there are no constraints. For searching with constraints, we set the minimal accuracy being 95.8\% ($\theta\geq \theta_M=95.8\%$) as an example; we compare FLASH with one-level SHGO because RL does not converge. The quality of the model is calculated by the objective function in Equation~\ref{eq:problem_definition} (higher is better). 
}
\scalebox{0.88}{
\begin{tabular}{|p{1.7cm}|p{4.4cm}|p{1.6cm}|p{1.6cm}|p{2.8cm}|l|l|}
\hline
                Constraints involved?  &  Method & Search cost (\#Samples) & Search Time (s)& Quality of obtained model (Eq.~\ref{eq:problem_definition}) & Converge?   \\ \hline\hline
\multirow{4}{*}{No} & RL & 10000 & 1955 & 20984 & Yes   \\ \cline{2-6} 
                  &  one-level SHGO & 23 & 0.03 & 20984 & Yes   \\ \cline{2-6} 
                  & \textbf{hierarchical SHGO (FLASH)}  & 69 & 0.07 & 20984 &  Yes  \\ \cline{2-6}
                  & \textbf{Improvement}  & 144.93$\times$ & 27929$\times$  & $1\times$ &  -  \\ \hline\hline
\multirow{4}{*}{Yes. $\theta\geq \theta_M$} & RL & >10000 &  -& - &  No  \\ \cline{2-6} 
                  & one-level SHGO  & 1195 & 3.82 & 10550 &  Yes  \\ \cline{2-6} 
                  & \textbf{hierarchical SHGO (FLASH)} &170  & 0.26 &  11969&  Yes   \\ \cline{2-6} 
                  & \textbf{Improvement} & 7.03$\times$ & 14.7$\times$ &  1.13$\times$&  - \\ \hline
\end{tabular}}
\label{tab:sea_alg_comp}
\end{table}

We compare the baseline-RL approach with our proposed SHGO-based optimization approach.
As shown in Table \ref{tab:sea_alg_comp}, when there is no constraint in terms of accuracy and hardware performance, our hierarchical SHGO-based algorithm brings negligible overhead compared to the one-level SHGO algorithm. Moreover, our hierarchical SHGO-based algorithm needs much fewer samples ($144.93\times$) during the search process than RL-based methods. Our proposed search algorithm is as fast as 0.07 seconds and 27929$\times$ faster than the RL-based methods, while achieving the same quality of the solution! As for the searching with specific constraints, the training of RL-based methods cannot even converge after training with 10000 samples. Furthermore, our hierarchical SHGO-based algorithm obtains a better-quality model with $7.03\times$ fewer samples and $14.7\times$ less search time compared to the one-level SHGO algorithm. The results show that our proposed hierarchical strategy further improves the efficiency of the original SHGO algorithm. 

\begin{figure} [t]
    \centering
    \vspace{0mm}
    \includegraphics[width=0.88\textwidth]{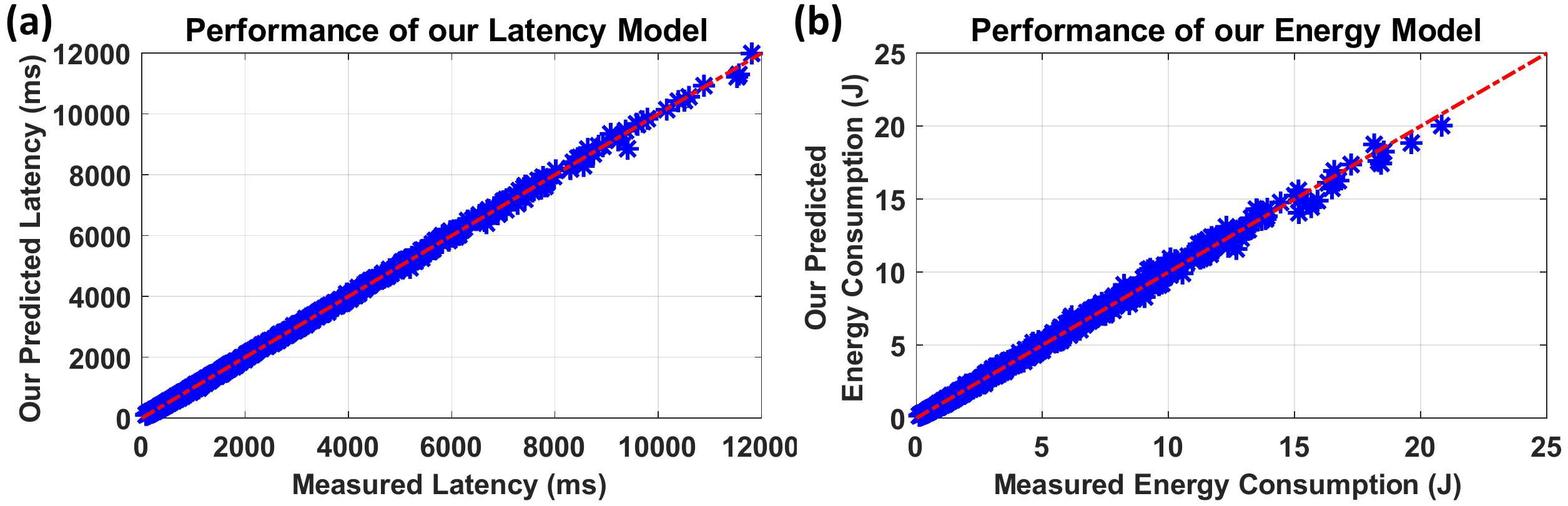}
    \vspace{-4mm}
    \caption{(a) Predictions of our analytical latency models vs. measured values for RPi-3B. (b) Predictions of our analytical energy consumption models vs. measured values for RPi-3B. The red dotted lines in these two figures show a high correlation between predicted and measured values.}
    \label{fig:rpi_hw_model_perf}
\end{figure}

\begin{figure} [ht]
    \centering
    \includegraphics[width=0.88\textwidth]{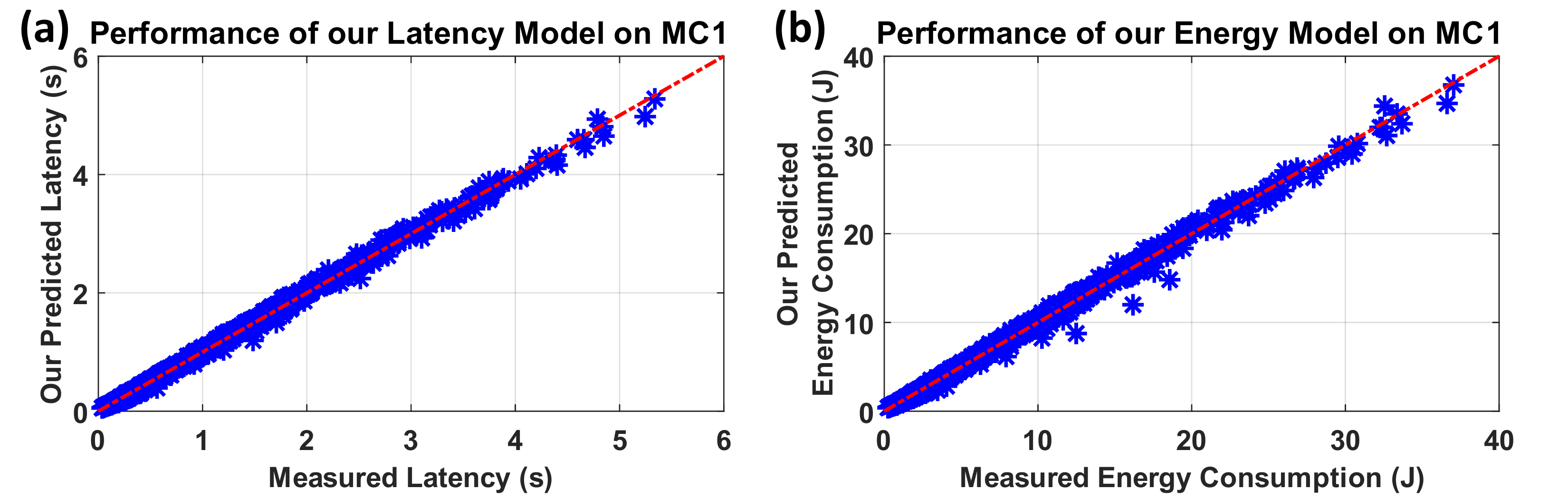}
    \vspace{-4mm}
    \caption{(a) Predictions of our analytical latency models vs. measured values for MC1. (b) Predictions of our analytical energy consumption models vs. measured values for MC1. The red dotted lines in these two figures show a very good correlation between the predicted and measured values.}
    \label{fig:mc1_hw_model_perf}
\end{figure}


\begin{table}[b]
 \caption{Comparison between one-level and hierarchical SHGO-based search on RPi-3B and Odroid MC1. For searching with constraints, we set the minimal accuracy being 96\% ($\theta\geq \theta_M=96\%$) as an example. The quality of the model is calculated by Equation~\ref{eq:rasp_obj} (higher is better).}
 \vspace{-4mm}
  \scalebox{0.88}{
\begin{tabular}{|l|l|l|l|l|l|l|l|}
\hline
\multirow{2}{*}{\begin{tabular}[c]{@{}l@{}}Constraints\\ involved?\end{tabular}} & \multirow{2}{*}{Method}   & \multicolumn{2}{l|}{\begin{tabular}[c]{@{}l@{}}Search Cost\\ (\# Samples)\end{tabular}} & \multicolumn{2}{l|}{\begin{tabular}[c]{@{}l@{}}Search time\\ (s)\end{tabular}} & \multicolumn{2}{l|}{\begin{tabular}[c]{@{}l@{}}Model Quality\\ (Equation~\ref{eq:rasp_obj})\end{tabular}} \\ \cline{3-8} 
                                                                                 &                  & RPi-3B                                             & MC1                                   & RPi-3B                                            & MC1                                 & RPi-3B                                            & MC1                                  \\ \hline \hline
\multirow{2}{*}{No}                                                              & one-level SHGO            & 112                                             &         113                              & 1.68                                           &       0.71                              & 4.74                                           &       4.13                               \\ \cline{2-8} 
                                                                                 & hierarchical SHGO (FLASH) & 180                                             & 135                                      & 2.21                                           & 0.45                                    & 4.74                                           &                                      4.13 \\ \hline\hline
\multirow{3}{*}{Yes, $\theta \geq \theta_M$}                                     & one-level SHGO            & 1309                                            &           1272                            & 45.98                                          &       9.65                              & 0.35                                            &    0.38                                  \\ \cline{2-8} 
                                                                                 & hierarchical SHGO (FLASH) & 261                                             &  414                                     & 2.33                                           &   1.32                                  & 0.48                                           &                                      0.57 \\ \cline{2-8} 
                                                                                 & \textbf{Improvement}               & \textbf{5.01 $\times$}                                   &            \textbf{3.07 $\times$}    & \textbf{19.73 $\times$}                                 &                   \textbf{20.5 $\times$}  & \textbf{1.37 $\times$ }                                 &  \textbf{1.51 $\times$}\\ \hline
\end{tabular}}
\label{tab:rpi_sea_alg}
\vspace{-5mm}
\end{table}

\vspace{-1mm}
\subsection{Case study: Raspberry Pi and Odroid MC1}
\vspace{-1mm}
As discussed in previous sections, each component and stages of FLASH are very efficient in terms of both computation and time costs. To further demonstrate the efficiency of our FLASH methodology, we implement FLASH on two typical edge devices, namely, the Raspberry Pi-3 Model-B (RPi-3B) and Odroid MC1 (MC1). 

\noindent\textbf{Setup:} RPi-3B has an Arm Cortex-A53 quad-core processor with a nominal frequency of 1.2GHz and 1GB of RAM. Furthermore, we use the Odroid Smart Power 2 to measure voltage, current, and power. We use TensorFlow-Lite (TF-Lite) as the run-time framework on RPi-3B. To achieve this, we first define the architecture of the models by TensorFlow (TF). Then we convert the TF model into the TF-Lite format and generate the binary file deployed on the RPi-3B. 

Odroid MC1 is powered by Exynos 5422, a heterogeneous system-on-a-chip (MPSoC). This SoC consists of two clusters of ARM cores and a small GPU core. Besides the hardware platform itself, we use the same setup as for the RPi-3B.

\noindent\textbf{Accuracy predictor and analytical hardware performance models:} We adopt the same accuracy predictor used in Section \ref{sec:res_nas}. We only consider latency and energy consumption as the hardware performance metrics because the chip area is fixed. Hence, the objective function of searching on RPi-3B and MC1 is:
\begin{equation} \label{eq:rasp_obj}
   f_{obj}=\frac{Accuracy}{Latency \times Energy} 
\end{equation}

To fine-tune the analytical latency and energy models, we randomly select 180 sample networks from the search space. Then we convert them into the TF-Lite format and record their latency and energy consumption on the RPi-3B. Based on the recorded data, we update the parameters of the analytical latency and energy models. Fig. \ref{fig:rpi_hw_model_perf} and \ref{fig:mc1_hw_model_perf} show that our analytical hardware performance models almost coincide with the real performance of both the RPi-3B and MC1.

\noindent\textbf{Search Process on RPi-3B and MC1:} We do not show the results of RL-based methods because the training of RL models requires intensive computation resources; thus, they cannot be deployed on RPi-3B and MC1. As shown in Table \ref{tab:rpi_sea_alg}, for searching without any constraint, our hierarchical SHGO-based algorithm has only a minimal overhead compared with the basic (one-level) SHGO algorithm. Moreover, our hierarchical SHGO-based algorithm is faster than the one-level SHGO algorithm on MC1.

For searching with constraints, the hierarchical SHGO-based algorithm obtains a better-quality model with $5.01\times$ fewer samples and $19.73\times$ less search time on the RPi-3B; we achieve similar improvements on MC1 as well. These results prove the effectiveness of our hierarchical strategy again. Overall, the total searching time on RPi-3B and MC1 are as short as 2.33 seconds and 1.32 seconds, respectively on such resource-constrained edge devices.
To our best knowledge, this is the first time when a neural architecture search is reported on edge devices.

%% file: 5-conclusion.tex
\vspace{-2mm}
\section{Conclusions and Future Work} 
\label{sec:conclusion}
\vspace{-1mm}

This paper presented a very fast methodology, called FLASH, to improve the time efficiency of NAS. To this end, we have proposed a new topology-based metric, namely the \textit{NN-Degree}. Using the NN-Degree, we have proposed an analytical accuracy predictor by training as few as 25 samples out of a vast search space with more than 63 billion configurations. 
Our proposed accuracy predictor achieves the same performance with 6.88$\times$ fewer samples and $65.79\times$ reduction in fine-tuning time cost compared to state-of-the-art approaches.
We have also optimized the on-chip communication by designing a mesh-NoC for communication across multiple layers; based on the optimized hardware, we have built new analytical models to predict area, latency, and energy consumption.

Combining the accuracy predictor and the analytical hardware performance models, we have developed a hierarchical simplicial homology global optimization (SHGO)-based algorithm to optimize the co-design process while considering both test accuracy and the area, latency, and energy figures of the target hardware. Finally, we have demonstrated that our newly proposed hierarchical SHGO-based algorithm enables 27729$\times$ faster (less than 0.1 seconds) NAS compared to the state-of-the-art RL-based approaches.
We have also shown that FLASH can be readily transferred to other hardware platforms by doing NAS on a Raspberry Pi-3B and Odroid MC1 in less than 3 seconds. To our best knowledge, our work is the first to report NAS performed directly and efficiently on edge devices.

We note that there is no fundamental limitation to apply FLASH to other machine learning tasks. However, no IMC-based architectures are widely adopted yet for other machine learning tasks like speech recognition or object segmentation. Therefore,the current work focuses on DNN inference and leaves the extension to other machine learning tasks as future work. Finally, we plan to incorporate more types of networks such as ResNet and MobileNet-v2 as part of our future work.
\vspace{-2mm}


%% file: main.bbl

\begin{thebibliography}{61}


\ifx \showCODEN    \undefined \def \showCODEN     #1{\unskip}     \fi
\ifx \showDOI      \undefined \def \showDOI       #1{#1}\fi
\ifx \showISBNx    \undefined \def \showISBNx     #1{\unskip}     \fi
\ifx \showISBNxiii \undefined \def \showISBNxiii  #1{\unskip}     \fi
\ifx \showISSN     \undefined \def \showISSN      #1{\unskip}     \fi
\ifx \showLCCN     \undefined \def \showLCCN      #1{\unskip}     \fi
\ifx \shownote     \undefined \def \shownote      #1{#1}          \fi
\ifx \showarticletitle \undefined \def \showarticletitle #1{#1}   \fi
\ifx \showURL      \undefined \def \showURL       {\relax}        \fi
\providecommand\bibfield[2]{#2}
\providecommand\bibinfo[2]{#2}
\providecommand\natexlab[1]{#1}
\providecommand\showeprint[2][]{arXiv:#2}

\bibitem[\protect\citeauthoryear{Abdelfattah, Mehrotra, Dudziak, and
  Lane}{Abdelfattah et~al\mbox{.}}{2021}]%
        {tf_nas1}
\bibfield{author}{\bibinfo{person}{Mohamed~S Abdelfattah},
  \bibinfo{person}{Abhinav Mehrotra}, \bibinfo{person}{{\L}ukasz Dudziak},
  {and} \bibinfo{person}{Nicholas~Donald Lane}.}
  \bibinfo{year}{2021}\natexlab{}.
\newblock \showarticletitle{Zero-Cost Proxies for Lightweight NAS}. In
  \bibinfo{booktitle}{\emph{International Conference on Learning
  Representations}}.
\newblock


\bibitem[\protect\citeauthoryear{Baker, Gupta, Naik, and Raskar}{Baker
  et~al\mbox{.}}{2016}]%
        {baker_17}
\bibfield{author}{\bibinfo{person}{Bowen Baker}, \bibinfo{person}{Otkrist
  Gupta}, \bibinfo{person}{Nikhil Naik}, {and} \bibinfo{person}{Ramesh
  Raskar}.} \bibinfo{year}{2016}\natexlab{}.
\newblock \showarticletitle{{Designing Neural Network Architectures using
  Reinforcement Learning}}.
\newblock \bibinfo{journal}{\emph{arXiv preprint arXiv:1611.02167}}
  (\bibinfo{year}{2016}).
\newblock


\bibitem[\protect\citeauthoryear{Barab{\'a}si and Bonabeau}{Barab{\'a}si and
  Bonabeau}{2003}]%
        {barabasi2003scale_Free}
\bibfield{author}{\bibinfo{person}{Albert-L{\'a}szl{\'o} Barab{\'a}si} {and}
  \bibinfo{person}{Eric Bonabeau}.} \bibinfo{year}{2003}\natexlab{}.
\newblock \showarticletitle{{Scale-free Networks}}.
\newblock \bibinfo{journal}{\emph{Scientific american}} \bibinfo{volume}{288},
  \bibinfo{number}{5} (\bibinfo{year}{2003}), \bibinfo{pages}{60--69}.
\newblock


\bibitem[\protect\citeauthoryear{Benmeziane et~al\mbox{.}}{Benmeziane
  et~al\mbox{.}}{2021}]%
        {benmeziane2021comprehensive}
\bibfield{author}{\bibinfo{person}{Hadjer Benmeziane} {et~al\mbox{.}}}
  \bibinfo{year}{2021}\natexlab{}.
\newblock \showarticletitle{A Comprehensive Survey on Hardware-Aware Neural
  Architecture Search}.
\newblock \bibinfo{journal}{\emph{arXiv preprint arXiv:2101.09336}}
  (\bibinfo{year}{2021}).
\newblock


\bibitem[\protect\citeauthoryear{Bhardwaj, Li, and Marculescu}{Bhardwaj
  et~al\mbox{.}}{2021}]%
        {nn_mass}
\bibfield{author}{\bibinfo{person}{Kartikeya Bhardwaj},
  \bibinfo{person}{Guihong Li}, {and} \bibinfo{person}{Radu Marculescu}.}
  \bibinfo{year}{2021}\natexlab{}.
\newblock \showarticletitle{How Does Topology Influence Gradient Propagation
  and Model Performance of Deep Networks With DenseNet-Type Skip Connections?}.
  In \bibinfo{booktitle}{\emph{Proceedings of the IEEE/CVF Conference on
  Computer Vision and Pattern Recognition (CVPR)}}.
\newblock


\bibitem[\protect\citeauthoryear{Brown et~al\mbox{.}}{Brown
  et~al\mbox{.}}{2020}]%
        {big_nlp}
\bibfield{author}{\bibinfo{person}{Tom~B Brown} {et~al\mbox{.}}}
  \bibinfo{year}{2020}\natexlab{}.
\newblock \showarticletitle{{Language Models are Few-Shot Learners}}.
\newblock \bibinfo{journal}{\emph{arXiv preprint arXiv:2005.14165}}
  (\bibinfo{year}{2020}).
\newblock


\bibitem[\protect\citeauthoryear{Cai, Gan, Wang, Zhang, and Han}{Cai
  et~al\mbox{.}}{2020}]%
        {Cai2020Once-for-All}
\bibfield{author}{\bibinfo{person}{Han Cai}, \bibinfo{person}{Chuang Gan},
  \bibinfo{person}{Tianzhe Wang}, \bibinfo{person}{Zhekai Zhang}, {and}
  \bibinfo{person}{Song Han}.} \bibinfo{year}{2020}\natexlab{}.
\newblock \showarticletitle{Once-for-All: Train One Network and Specialize it
  for Efficient Deployment}. In \bibinfo{booktitle}{\emph{International
  Conference on Learning Representations}}.
\newblock


\bibitem[\protect\citeauthoryear{Cai, Zhu, and Han}{Cai et~al\mbox{.}}{2019}]%
        {cai2018proxylessnas}
\bibfield{author}{\bibinfo{person}{Han Cai}, \bibinfo{person}{Ligeng Zhu},
  {and} \bibinfo{person}{Song Han}.} \bibinfo{year}{2019}\natexlab{}.
\newblock \showarticletitle{Proxyless{NAS}: Direct Neural Architecture Search
  on Target Task and Hardware}. In \bibinfo{booktitle}{\emph{International
  Conference on Learning Representations}}.
\newblock


\bibitem[\protect\citeauthoryear{Chau, Dudziak, Abdelfattah, Lee, Kim, and
  Lane}{Chau et~al\mbox{.}}{2020}]%
        {brp_nas}
\bibfield{author}{\bibinfo{person}{Thomas Chau}, \bibinfo{person}{{\L}ukasz
  Dudziak}, \bibinfo{person}{Mohamed~S Abdelfattah}, \bibinfo{person}{Royson
  Lee}, \bibinfo{person}{Hyeji Kim}, {and} \bibinfo{person}{Nicholas~D Lane}.}
  \bibinfo{year}{2020}\natexlab{}.
\newblock \showarticletitle{{BRP-NAS: Prediction-based NAS using GCNs}}.
\newblock \bibinfo{journal}{\emph{arXiv preprint arXiv:2007.08668}}
  (\bibinfo{year}{2020}).
\newblock


\bibitem[\protect\citeauthoryear{Chen, Peng, and Yu}{Chen
  et~al\mbox{.}}{2018}]%
        {chen2018neurosim}
\bibfield{author}{\bibinfo{person}{Pai-Yu Chen}, \bibinfo{person}{Xiaochen
  Peng}, {and} \bibinfo{person}{Shimeng Yu}.} \bibinfo{year}{2018}\natexlab{}.
\newblock \showarticletitle{{NeuroSim: A Circuit-level Macro Model for
  Benchmarking Neuro-Inspired Architectures in Online Learning}}.
\newblock \bibinfo{journal}{\emph{IEEE Transactions on Computer-Aided Design of
  Integrated Circuits and Systems}} \bibinfo{volume}{37}, \bibinfo{number}{12}
  (\bibinfo{year}{2018}), \bibinfo{pages}{3067--3080}.
\newblock


\bibitem[\protect\citeauthoryear{Chen, Gong, and Wang}{Chen
  et~al\mbox{.}}{2021}]%
        {tf_nas2}
\bibfield{author}{\bibinfo{person}{Wuyang Chen}, \bibinfo{person}{Xinyu Gong},
  {and} \bibinfo{person}{Zhangyang Wang}.} \bibinfo{year}{2021}\natexlab{}.
\newblock \showarticletitle{Neural Architecture Search on ImageNet in Four GPU
  Hours: A Theoretically Inspired Perspective}. In
  \bibinfo{booktitle}{\emph{International Conference on Learning
  Representations}}.
\newblock


\bibitem[\protect\citeauthoryear{Chiang et~al\mbox{.}}{Chiang
  et~al\mbox{.}}{2019}]%
        {chiang2019cluster}
\bibfield{author}{\bibinfo{person}{Wei-Lin Chiang} {et~al\mbox{.}}}
  \bibinfo{year}{2019}\natexlab{}.
\newblock \showarticletitle{{Cluster-gcn: An Efficient Algorithm for Training
  Deep and Large Graph Convolutional Networks}}. In
  \bibinfo{booktitle}{\emph{Proceedings of the 25th ACM SIGKDD International
  Conference on Knowledge Discovery \& Data Mining}}.
  \bibinfo{pages}{257--266}.
\newblock


\bibitem[\protect\citeauthoryear{Courbariaux, Hubara, Soudry, El-Yaniv, and
  Bengio}{Courbariaux et~al\mbox{.}}{2016}]%
        {bnn}
\bibfield{author}{\bibinfo{person}{Matthieu Courbariaux}, \bibinfo{person}{Itay
  Hubara}, \bibinfo{person}{Daniel Soudry}, \bibinfo{person}{Ran El-Yaniv},
  {and} \bibinfo{person}{Yoshua Bengio}.} \bibinfo{year}{2016}\natexlab{}.
\newblock \showarticletitle{{Binarized Neural Networks: Training Deep Neural
  Networks with Weights and Activations Constrained to+ 1 or-1}}.
\newblock \bibinfo{journal}{\emph{arXiv preprint arXiv:1602.02830}}
  (\bibinfo{year}{2016}).
\newblock


\bibitem[\protect\citeauthoryear{Dai, Yin, and Jha}{Dai et~al\mbox{.}}{2020}]%
        {jha_dac20}
\bibfield{author}{\bibinfo{person}{Xiaoliang Dai}, \bibinfo{person}{Hongxu
  Yin}, {and} \bibinfo{person}{Niraj~K. Jha}.} \bibinfo{year}{2020}\natexlab{}.
\newblock \showarticletitle{Grow and Prune Compact, Fast, and Accurate LSTMs}.
\newblock \bibinfo{journal}{\emph{IEEE Trans. Comput.}} \bibinfo{volume}{69},
  \bibinfo{number}{3} (\bibinfo{year}{2020}), \bibinfo{pages}{441--452}.
\newblock


\bibitem[\protect\citeauthoryear{Deng, Dong, Socher, Li, Li, and Fei-Fei}{Deng
  et~al\mbox{.}}{2009}]%
        {img_net}
\bibfield{author}{\bibinfo{person}{Jia Deng}, \bibinfo{person}{Wei Dong},
  \bibinfo{person}{Richard Socher}, \bibinfo{person}{Li-Jia Li},
  \bibinfo{person}{Kai Li}, {and} \bibinfo{person}{Li Fei-Fei}.}
  \bibinfo{year}{2009}\natexlab{}.
\newblock \showarticletitle{{Imagenet: A Large-Scale Hierarchical Image
  Database}}. In \bibinfo{booktitle}{\emph{Proceedings of the IEEE Conference
  on Computer Vision and Pattern Recognition (CVPR)}}.
  \bibinfo{pages}{248--255}.
\newblock


\bibitem[\protect\citeauthoryear{Dong and Yang}{Dong and Yang}{2020}]%
        {dong2020nasbench201}
\bibfield{author}{\bibinfo{person}{Xuanyi Dong} {and} \bibinfo{person}{Yi
  Yang}.} \bibinfo{year}{2020}\natexlab{}.
\newblock \showarticletitle{{NAS-Bench-201: Extending the Scope of Reproducible
  Neural Architecture Search}}.
\newblock \bibinfo{journal}{\emph{arXiv preprint arXiv:2001.00326}}
  (\bibinfo{year}{2020}).
\newblock


\bibitem[\protect\citeauthoryear{Elsken, Metzen, and Hutter}{Elsken
  et~al\mbox{.}}{2019}]%
        {elsken2019neural}
\bibfield{author}{\bibinfo{person}{Thomas Elsken}, \bibinfo{person}{Jan~Hendrik
  Metzen}, {and} \bibinfo{person}{Frank Hutter}.}
  \bibinfo{year}{2019}\natexlab{}.
\newblock \showarticletitle{Neural architecture search: A survey}.
\newblock \bibinfo{journal}{\emph{The Journal of Machine Learning Research}}
  \bibinfo{volume}{20}, \bibinfo{number}{1} (\bibinfo{year}{2019}),
  \bibinfo{pages}{1997--2017}.
\newblock


\bibitem[\protect\citeauthoryear{Endres, Sandrock, and Focke}{Endres
  et~al\mbox{.}}{2018}]%
        {shgo}
\bibfield{author}{\bibinfo{person}{Stefan~C Endres}, \bibinfo{person}{Carl
  Sandrock}, {and} \bibinfo{person}{Walter~W Focke}.}
  \bibinfo{year}{2018}\natexlab{}.
\newblock \showarticletitle{{A Simplicial Homology Algorithm for Lipschitz
  Optimisation}}.
\newblock \bibinfo{journal}{\emph{Journal of Global Optimization}}
  \bibinfo{volume}{72}, \bibinfo{number}{2} (\bibinfo{year}{2018}),
  \bibinfo{pages}{181--217}.
\newblock


\bibitem[\protect\citeauthoryear{Grot and Keckler}{Grot and Keckler}{2008}]%
        {grot2008scalable}
\bibfield{author}{\bibinfo{person}{Boris Grot} {and} \bibinfo{person}{Stephen~W
  Keckler}.} \bibinfo{year}{2008}\natexlab{}.
\newblock \showarticletitle{{Scalable On-Chip Interconnect Topologies}}. In
  \bibinfo{booktitle}{\emph{2nd Workshop on Chip Multiprocessor Memory Systems
  and Interconnects}}.
\newblock


\bibitem[\protect\citeauthoryear{Han, Mao, and Dally}{Han
  et~al\mbox{.}}{2015}]%
        {han_prune}
\bibfield{author}{\bibinfo{person}{Song Han}, \bibinfo{person}{Huizi Mao},
  {and} \bibinfo{person}{William~J Dally}.} \bibinfo{year}{2015}\natexlab{}.
\newblock \showarticletitle{{Deep Compression: Compressing Deep Neural Networks
  with Pruning, Trained Quantization and Huffman Coding}}.
\newblock \bibinfo{journal}{\emph{arXiv preprint arXiv:1510.00149}}
  (\bibinfo{year}{2015}).
\newblock


\bibitem[\protect\citeauthoryear{He, Zhang, Ren, and Sun}{He
  et~al\mbox{.}}{2016}]%
        {resnet}
\bibfield{author}{\bibinfo{person}{Kaiming He}, \bibinfo{person}{Xiangyu
  Zhang}, \bibinfo{person}{Shaoqing Ren}, {and} \bibinfo{person}{Jian Sun}.}
  \bibinfo{year}{2016}\natexlab{}.
\newblock \showarticletitle{{Deep Residual Learning for Image Recognition}}. In
  \bibinfo{booktitle}{\emph{Proceedings of the IEEE conference on computer
  vision and pattern recognition}}. \bibinfo{pages}{770--778}.
\newblock


\bibitem[\protect\citeauthoryear{Hinton, Vinyals, and Dean}{Hinton
  et~al\mbox{.}}{2015}]%
        {kd}
\bibfield{author}{\bibinfo{person}{Geoffrey Hinton}, \bibinfo{person}{Oriol
  Vinyals}, {and} \bibinfo{person}{Jeff Dean}.}
  \bibinfo{year}{2015}\natexlab{}.
\newblock \showarticletitle{{Distilling the Knowledge in a Neural Network}}.
\newblock \bibinfo{journal}{\emph{arXiv preprint arXiv:1503.02531}}
  (\bibinfo{year}{2015}).
\newblock


\bibitem[\protect\citeauthoryear{Horowitz}{Horowitz}{2014}]%
        {horowitz20141}
\bibfield{author}{\bibinfo{person}{Mark Horowitz}.}
  \bibinfo{year}{2014}\natexlab{}.
\newblock \showarticletitle{{1.1 Computing's Energy Problem (and what we can do
  about it)}}. In \bibinfo{booktitle}{\emph{IEEE International Solid-State
  Circuits Conference Digest of Technical Papers (ISSCC)}}.
  \bibinfo{pages}{10--14}.
\newblock


\bibitem[\protect\citeauthoryear{Hsu et~al\mbox{.}}{Hsu et~al\mbox{.}}{2018}]%
        {hsu2018monas}
\bibfield{author}{\bibinfo{person}{Chi-Hung Hsu} {et~al\mbox{.}}}
  \bibinfo{year}{2018}\natexlab{}.
\newblock \showarticletitle{{Monas: Multi-objective Neural Architecture Search
  using Reinforcement Learning}}.
\newblock \bibinfo{journal}{\emph{arXiv preprint arXiv:1806.10332}}
  (\bibinfo{year}{2018}).
\newblock


\bibitem[\protect\citeauthoryear{Huang, Liu, Van Der~Maaten, and
  Weinberger}{Huang et~al\mbox{.}}{2017}]%
        {densenet}
\bibfield{author}{\bibinfo{person}{Gao Huang}, \bibinfo{person}{Zhuang Liu},
  \bibinfo{person}{Laurens Van Der~Maaten}, {and} \bibinfo{person}{Kilian~Q
  Weinberger}.} \bibinfo{year}{2017}\natexlab{}.
\newblock \showarticletitle{{Densely Connected Convolutional Networks}}. In
  \bibinfo{booktitle}{\emph{Proceedings of the IEEE conference on computer
  vision and pattern recognition}}. \bibinfo{pages}{4700--4708}.
\newblock


\bibitem[\protect\citeauthoryear{Jiang et~al\mbox{.}}{Jiang
  et~al\mbox{.}}{2013}]%
        {jiang2013detailed}
\bibfield{author}{\bibinfo{person}{Nan Jiang} {et~al\mbox{.}}}
  \bibinfo{year}{2013}\natexlab{}.
\newblock \showarticletitle{{A Detailed and Flexible Cycle-Accurate
  Network-on-Chip Simulator}}. In \bibinfo{booktitle}{\emph{IEEE ISPASS}}.
  \bibinfo{pages}{86--96}.
\newblock


\bibitem[\protect\citeauthoryear{Jiang et~al\mbox{.}}{Jiang
  et~al\mbox{.}}{2020}]%
        {jiang2020device}
\bibfield{author}{\bibinfo{person}{Weiwen Jiang} {et~al\mbox{.}}}
  \bibinfo{year}{2020}\natexlab{}.
\newblock \showarticletitle{{Device-circuit-architecture Co-exploration for
  Computing-in-memory Neural Accelerators}}.
\newblock \bibinfo{journal}{\emph{IEEE Trans. Comput.}} (\bibinfo{year}{2020}).
\newblock


\bibitem[\protect\citeauthoryear{Krishnan et~al\mbox{.}}{Krishnan
  et~al\mbox{.}}{2020}]%
        {krishnan2020interconnect}
\bibfield{author}{\bibinfo{person}{Gokul Krishnan} {et~al\mbox{.}}}
  \bibinfo{year}{2020}\natexlab{}.
\newblock \showarticletitle{{Interconnect-aware Area and Energy Optimization
  for In-memory Acceleration of DNNs}}.
\newblock \bibinfo{journal}{\emph{IEEE Design \& Test}} \bibinfo{volume}{37},
  \bibinfo{number}{6} (\bibinfo{year}{2020}), \bibinfo{pages}{79--87}.
\newblock


\bibitem[\protect\citeauthoryear{Krishnan et~al\mbox{.}}{Krishnan
  et~al\mbox{.}}{2021}]%
        {krishnan2021interconnect}
\bibfield{author}{\bibinfo{person}{Gokul Krishnan} {et~al\mbox{.}}}
  \bibinfo{year}{2021}\natexlab{}.
\newblock \showarticletitle{{Interconnect-Centric Benchmarking of In-Memory
  Acceleration for DNNS}}. In \bibinfo{booktitle}{\emph{2021 China
  Semiconductor Technology International Conference (CSTIC)}}. IEEE,
  \bibinfo{pages}{1--4}.
\newblock


\bibitem[\protect\citeauthoryear{Li et~al\mbox{.}}{Li et~al\mbox{.}}{2020}]%
        {edd_Dac}
\bibfield{author}{\bibinfo{person}{Yuhong Li} {et~al\mbox{.}}}
  \bibinfo{year}{2020}\natexlab{}.
\newblock \showarticletitle{EDD: Efficient Differentiable DNN Architecture and
  Implementation Co-Search for Embedded AI Solutions}. In
  \bibinfo{booktitle}{\emph{Proceedings of the 57th ACM/EDAC/IEEE Design
  Automation Conference}}. \bibinfo{publisher}{IEEE Press}, Article
  \bibinfo{articleno}{130}, \bibinfo{numpages}{6}~pages.
\newblock


\bibitem[\protect\citeauthoryear{Liu et~al\mbox{.}}{Liu et~al\mbox{.}}{2018a}]%
        {lstm}
\bibfield{author}{\bibinfo{person}{Chenxi Liu} {et~al\mbox{.}}}
  \bibinfo{year}{2018}\natexlab{a}.
\newblock \showarticletitle{{Progressive Neural Architecture Search}}. In
  \bibinfo{booktitle}{\emph{Proceedings of the European Conference on Computer
  Vision (ECCV)}}. \bibinfo{pages}{19--34}.
\newblock


\bibitem[\protect\citeauthoryear{Liu, Simonyan, and Yang}{Liu
  et~al\mbox{.}}{2018b}]%
        {Darts}
\bibfield{author}{\bibinfo{person}{Hanxiao Liu}, \bibinfo{person}{Karen
  Simonyan}, {and} \bibinfo{person}{Yiming Yang}.}
  \bibinfo{year}{2018}\natexlab{b}.
\newblock \showarticletitle{Darts: Differentiable Architecture Search}.
\newblock \bibinfo{journal}{\emph{arXiv preprint arXiv:1806.09055}}
  (\bibinfo{year}{2018}).
\newblock


\bibitem[\protect\citeauthoryear{Lukasik, Friede, Stuckenschmidt, and
  Keuper}{Lukasik et~al\mbox{.}}{2020}]%
        {pr_2020_gnn_acc_pre}
\bibfield{author}{\bibinfo{person}{Jovita Lukasik}, \bibinfo{person}{David
  Friede}, \bibinfo{person}{Heiner Stuckenschmidt}, {and}
  \bibinfo{person}{Margret Keuper}.} \bibinfo{year}{2020}\natexlab{}.
\newblock \showarticletitle{{Neural Architecture Performance Prediction Using
  Graph Neural Networks}}.
\newblock \bibinfo{journal}{\emph{arXiv preprint arXiv:2010.10024}}
  (\bibinfo{year}{2020}).
\newblock


\bibitem[\protect\citeauthoryear{Mandal et~al\mbox{.}}{Mandal
  et~al\mbox{.}}{2020}]%
        {mandal2020latency}
\bibfield{author}{\bibinfo{person}{Sumit~K Mandal} {et~al\mbox{.}}}
  \bibinfo{year}{2020}\natexlab{}.
\newblock \showarticletitle{{A Latency-Optimized Reconfigurable NoC for
  In-Memory Acceleration of DNNs}}.
\newblock \bibinfo{journal}{\emph{IEEE Journal on Emerging and Selected Topics
  in Circuits and Systems}} \bibinfo{volume}{10}, \bibinfo{number}{3}
  (\bibinfo{year}{2020}), \bibinfo{pages}{362--375}.
\newblock


\bibitem[\protect\citeauthoryear{Manning, Manning, and Sch{\"u}tze}{Manning
  et~al\mbox{.}}{1999}]%
        {manning1999foundations}
\bibfield{author}{\bibinfo{person}{Christopher~D Manning},
  \bibinfo{person}{Christopher~D Manning}, {and} \bibinfo{person}{Hinrich
  Sch{\"u}tze}.} \bibinfo{year}{1999}\natexlab{}.
\newblock \bibinfo{booktitle}{\emph{{Foundations of Statistical Natural
  Language Processing}}}.
\newblock \bibinfo{publisher}{MIT press}.
\newblock


\bibitem[\protect\citeauthoryear{Mao, Schwarzkopf, Venkatakrishnan, Meng, and
  Alizadeh}{Mao et~al\mbox{.}}{2019}]%
        {mao2019learning}
\bibfield{author}{\bibinfo{person}{Hongzi Mao}, \bibinfo{person}{Malte
  Schwarzkopf}, \bibinfo{person}{Shaileshh~Bojja Venkatakrishnan},
  \bibinfo{person}{Zili Meng}, {and} \bibinfo{person}{Mohammad Alizadeh}.}
  \bibinfo{year}{2019}\natexlab{}.
\newblock \showarticletitle{{Learning Scheduling Algorithms for Data Processing
  Clusters}}.
\newblock In \bibinfo{booktitle}{\emph{ACM Special Interest Group on Data
  Communication}}. \bibinfo{pages}{270--288}.
\newblock


\bibitem[\protect\citeauthoryear{Marculescu, Stamoulis, and Cai}{Marculescu
  et~al\mbox{.}}{2018}]%
        {diana_modeling}
\bibfield{author}{\bibinfo{person}{Diana Marculescu},
  \bibinfo{person}{Dimitrios Stamoulis}, {and} \bibinfo{person}{Ermao Cai}.}
  \bibinfo{year}{2018}\natexlab{}.
\newblock \showarticletitle{Hardware-Aware Machine Learning: Modeling and
  Optimization}. In \bibinfo{booktitle}{\emph{Proceedings of the International
  Conference on Computer-Aided Design}} \emph{(\bibinfo{series}{ICCAD '18})}.
\newblock


\bibitem[\protect\citeauthoryear{Mnih et~al\mbox{.}}{Mnih
  et~al\mbox{.}}{2013}]%
        {mnih2013playing}
\bibfield{author}{\bibinfo{person}{Volodymyr Mnih} {et~al\mbox{.}}}
  \bibinfo{year}{2013}\natexlab{}.
\newblock \showarticletitle{{Playing Atari with Deep Reinforcement Learning}}.
\newblock \bibinfo{journal}{\emph{arXiv preprint arXiv:1312.5602}}
  (\bibinfo{year}{2013}).
\newblock


\bibitem[\protect\citeauthoryear{Newman, Barab{\'a}si, and Watts}{Newman
  et~al\mbox{.}}{2006}]%
        {newman2006structure}
\bibfield{author}{\bibinfo{person}{Mark Newman},
  \bibinfo{person}{Albert-L{\'a}szl{\'o} Barab{\'a}si}, {and}
  \bibinfo{person}{Duncan~J Watts}.} \bibinfo{year}{2006}\natexlab{}.
\newblock \bibinfo{booktitle}{\emph{{The Structure and Dynamics of Networks.}}}
\newblock \bibinfo{publisher}{{Princeton University Press}}.
\newblock


\bibitem[\protect\citeauthoryear{Ning, Zheng, Zhao, Wang, and Yang}{Ning
  et~al\mbox{.}}{2020}]%
        {eccv_gates}
\bibfield{author}{\bibinfo{person}{Xuefei Ning}, \bibinfo{person}{Yin Zheng},
  \bibinfo{person}{Tianchen Zhao}, \bibinfo{person}{Yu Wang}, {and}
  \bibinfo{person}{Huazhong Yang}.} \bibinfo{year}{2020}\natexlab{}.
\newblock \showarticletitle{{A Generic Graph-based Neural Architecture Encoding
  Scheme for Predictor-based NAS}}.
\newblock  (\bibinfo{year}{2020}).
\newblock


\bibitem[\protect\citeauthoryear{Peng et~al\mbox{.}}{Peng
  et~al\mbox{.}}{2019}]%
        {peng2019inference}
\bibfield{author}{\bibinfo{person}{Xiaochen Peng} {et~al\mbox{.}}}
  \bibinfo{year}{2019}\natexlab{}.
\newblock \showarticletitle{{Inference Engine Benchmarking Across Technological
  Platforms from CMOS to RRAM}}. In \bibinfo{booktitle}{\emph{Proceedings of
  the International Symposium on Memory Systems}}. \bibinfo{pages}{471--479}.
\newblock


\bibitem[\protect\citeauthoryear{Pham, Guan, Zoph, Le, and Dean}{Pham
  et~al\mbox{.}}{2018}]%
        {hyper_nas}
\bibfield{author}{\bibinfo{person}{Hieu Pham}, \bibinfo{person}{Melody Guan},
  \bibinfo{person}{Barret Zoph}, \bibinfo{person}{Quoc Le}, {and}
  \bibinfo{person}{Jeff Dean}.} \bibinfo{year}{2018}\natexlab{}.
\newblock \showarticletitle{{Efficient Neural Architecture Search via
  Parameters Sharing}}. In \bibinfo{booktitle}{\emph{International Conference
  on Machine Learning}}. PMLR, \bibinfo{pages}{4095--4104}.
\newblock


\bibitem[\protect\citeauthoryear{Qiao, Cao, Yang, Song, and Li}{Qiao
  et~al\mbox{.}}{2018}]%
        {qiao2018atomlayer}
\bibfield{author}{\bibinfo{person}{Ximing Qiao}, \bibinfo{person}{Xiong Cao},
  \bibinfo{person}{Huanrui Yang}, \bibinfo{person}{Linghao Song}, {and}
  \bibinfo{person}{Hai Li}.} \bibinfo{year}{2018}\natexlab{}.
\newblock \showarticletitle{{Atomlayer: A Universal Reram-based CNN Accelerator
  with Atomic Layer Computation}}. In \bibinfo{booktitle}{\emph{IEEE/ACM DAC}}.
\newblock


\bibitem[\protect\citeauthoryear{Real et~al\mbox{.}}{Real
  et~al\mbox{.}}{2017}]%
        {real_17}
\bibfield{author}{\bibinfo{person}{Esteban Real} {et~al\mbox{.}}}
  \bibinfo{year}{2017}\natexlab{}.
\newblock \showarticletitle{{Large-scale Evolution of Image Classifiers}}. In
  \bibinfo{booktitle}{\emph{International Conference on Machine Learning}}.
  PMLR, \bibinfo{pages}{2902--2911}.
\newblock


\bibitem[\protect\citeauthoryear{Sandler, Howard, Zhu, Zhmoginov, and
  Chen}{Sandler et~al\mbox{.}}{2018}]%
        {mobilenetv2}
\bibfield{author}{\bibinfo{person}{Mark Sandler}, \bibinfo{person}{Andrew
  Howard}, \bibinfo{person}{Menglong Zhu}, \bibinfo{person}{Andrey Zhmoginov},
  {and} \bibinfo{person}{Liang-Chieh Chen}.} \bibinfo{year}{2018}\natexlab{}.
\newblock \showarticletitle{{Mobilenetv2: Inverted Residuals and Linear
  Bottlenecks}}. In \bibinfo{booktitle}{\emph{Proceedings of the IEEE
  conference on computer vision and pattern recognition}}.
  \bibinfo{pages}{4510--4520}.
\newblock


\bibitem[\protect\citeauthoryear{Shafiee et~al\mbox{.}}{Shafiee
  et~al\mbox{.}}{2016}]%
        {shafiee2016isaac}
\bibfield{author}{\bibinfo{person}{Ali Shafiee} {et~al\mbox{.}}}
  \bibinfo{year}{2016}\natexlab{}.
\newblock \showarticletitle{{ISAAC: A Convolutional Neural Network Accelerator
  with in-situ Analog Arithmetic in Crossbars}}.
\newblock \bibinfo{journal}{\emph{ACM/IEEE ISCA}} (\bibinfo{year}{2016}).
\newblock


\bibitem[\protect\citeauthoryear{Simonyan and Zisserman}{Simonyan and
  Zisserman}{2014}]%
        {simonyan2014vgg}
\bibfield{author}{\bibinfo{person}{Karen Simonyan} {and}
  \bibinfo{person}{Andrew Zisserman}.} \bibinfo{year}{2014}\natexlab{}.
\newblock \showarticletitle{{Very Deep Convolutional Networks for Large-scale
  Image Recognition}}.
\newblock \bibinfo{journal}{\emph{arXiv preprint arXiv:1409.1556}}
  (\bibinfo{year}{2014}).
\newblock


\bibitem[\protect\citeauthoryear{Song, Qian, Li, and Chen}{Song
  et~al\mbox{.}}{2017}]%
        {song2017pipelayer}
\bibfield{author}{\bibinfo{person}{Linghao Song}, \bibinfo{person}{Xuehai
  Qian}, \bibinfo{person}{Hai Li}, {and} \bibinfo{person}{Yiran Chen}.}
  \bibinfo{year}{2017}\natexlab{}.
\newblock \showarticletitle{{Pipelayer: A Pipelined Reram-based Accelerator for
  Deep Learning}}. In \bibinfo{booktitle}{\emph{IEEE HPCA}}.
  \bibinfo{pages}{541--552}.
\newblock


\bibitem[\protect\citeauthoryear{Stamoulis et~al\mbox{.}}{Stamoulis
  et~al\mbox{.}}{2019}]%
        {single_path_nas}
\bibfield{author}{\bibinfo{person}{Dimitrios Stamoulis} {et~al\mbox{.}}}
  \bibinfo{year}{2019}\natexlab{}.
\newblock \showarticletitle{{Single-Path NAS: Designing Hardware-Efficient
  ConvNets in less than 4 Hours}}.
\newblock \bibinfo{journal}{\emph{arXiv preprint arXiv:1904.02877}}
  (\bibinfo{year}{2019}).
\newblock


\bibitem[\protect\citeauthoryear{Tan et~al\mbox{.}}{Tan et~al\mbox{.}}{2019}]%
        {global_1}
\bibfield{author}{\bibinfo{person}{Mingxing Tan} {et~al\mbox{.}}}
  \bibinfo{year}{2019}\natexlab{}.
\newblock \showarticletitle{Mnasnet: Platform-aware neural architecture search
  for mobile}. In \bibinfo{booktitle}{\emph{Proceedings of the IEEE Conference
  on Computer Vision and Pattern Recognition (CVPR)}}.
  \bibinfo{pages}{2820--2828}.
\newblock


\bibitem[\protect\citeauthoryear{Veit, Wilber, and Belongie}{Veit
  et~al\mbox{.}}{2016}]%
        {ensemble_resnet}
\bibfield{author}{\bibinfo{person}{Andreas Veit}, \bibinfo{person}{Michael
  Wilber}, {and} \bibinfo{person}{Serge Belongie}.}
  \bibinfo{year}{2016}\natexlab{}.
\newblock \showarticletitle{{Residual Networks Behave Like Ensembles of
  Relatively Shallow Networks}}.
\newblock \bibinfo{journal}{\emph{arXiv preprint arXiv:1605.06431}}
  (\bibinfo{year}{2016}).
\newblock


\bibitem[\protect\citeauthoryear{Wang, Pathania, and Mitra}{Wang
  et~al\mbox{.}}{2020}]%
        {wang2020neural}
\bibfield{author}{\bibinfo{person}{Siqi Wang}, \bibinfo{person}{Anuj Pathania},
  {and} \bibinfo{person}{Tulika Mitra}.} \bibinfo{year}{2020}\natexlab{}.
\newblock \showarticletitle{{Neural Network Inference on Mobile SoCs}}.
\newblock \bibinfo{journal}{\emph{IEEE Design \& Test}} \bibinfo{volume}{37},
  \bibinfo{number}{5} (\bibinfo{year}{2020}), \bibinfo{pages}{50--57}.
\newblock


\bibitem[\protect\citeauthoryear{Watts and Strogatz}{Watts and
  Strogatz}{1998}]%
        {smallworldness}
\bibfield{author}{\bibinfo{person}{Duncan~J Watts} {and}
  \bibinfo{person}{Steven~H Strogatz}.} \bibinfo{year}{1998}\natexlab{}.
\newblock \showarticletitle{{Collective Dynamics of
  ‘Small-World’Networks}}.
\newblock \bibinfo{journal}{\emph{Nature}} \bibinfo{volume}{393},
  \bibinfo{number}{6684} (\bibinfo{year}{1998}), \bibinfo{pages}{440--442}.
\newblock


\bibitem[\protect\citeauthoryear{Wen, Liu, Chen, Li, Bender, and
  Kindermans}{Wen et~al\mbox{.}}{2020}]%
        {yiran_gnn}
\bibfield{author}{\bibinfo{person}{Wei Wen}, \bibinfo{person}{Hanxiao Liu},
  \bibinfo{person}{Yiran Chen}, \bibinfo{person}{Hai Li},
  \bibinfo{person}{Gabriel Bender}, {and} \bibinfo{person}{Pieter-Jan
  Kindermans}.} \bibinfo{year}{2020}\natexlab{}.
\newblock \showarticletitle{{Neural Predictor for Neural Architecture Search}}.
  In \bibinfo{booktitle}{\emph{European Conference on Computer Vision}}.
  Springer, \bibinfo{pages}{660--676}.
\newblock


\bibitem[\protect\citeauthoryear{Wistuba, Rawat, and Pedapati}{Wistuba
  et~al\mbox{.}}{2019}]%
        {nas_survey}
\bibfield{author}{\bibinfo{person}{Martin Wistuba}, \bibinfo{person}{Ambrish
  Rawat}, {and} \bibinfo{person}{Tejaswini Pedapati}.}
  \bibinfo{year}{2019}\natexlab{}.
\newblock \showarticletitle{{A Survey on Neural Architecture Search}}.
\newblock \bibinfo{journal}{\emph{arXiv preprint arXiv:1905.01392}}
  (\bibinfo{year}{2019}).
\newblock


\bibitem[\protect\citeauthoryear{Wu et~al\mbox{.}}{Wu et~al\mbox{.}}{2019}]%
        {wu2019fbnet}
\bibfield{author}{\bibinfo{person}{Bichen Wu} {et~al\mbox{.}}}
  \bibinfo{year}{2019}\natexlab{}.
\newblock \showarticletitle{{Fbnet: Hardware-aware Efficient Convnet Design via
  Differentiable Neural Architecture Search}}. In
  \bibinfo{booktitle}{\emph{Proceedings of the IEEE/CVF Conference on Computer
  Vision and Pattern Recognition}}. \bibinfo{pages}{10734--10742}.
\newblock


\bibitem[\protect\citeauthoryear{Xie, Zheng, Liu, and Lin}{Xie
  et~al\mbox{.}}{2019}]%
        {xie2018snas}
\bibfield{author}{\bibinfo{person}{Sirui Xie}, \bibinfo{person}{Hehui Zheng},
  \bibinfo{person}{Chunxiao Liu}, {and} \bibinfo{person}{Liang Lin}.}
  \bibinfo{year}{2019}\natexlab{}.
\newblock \showarticletitle{{SNAS}: stochastic neural architecture search}. In
  \bibinfo{booktitle}{\emph{International Conference on Learning
  Representations}}.
\newblock


\bibitem[\protect\citeauthoryear{Yu et~al\mbox{.}}{Yu et~al\mbox{.}}{2020}]%
        {big_nas}
\bibfield{author}{\bibinfo{person}{Jiahui Yu} {et~al\mbox{.}}}
  \bibinfo{year}{2020}\natexlab{}.
\newblock \showarticletitle{BigNAS: Scaling up Neural Architecture Search with
  Big Single-Stage Models}. In \bibinfo{booktitle}{\emph{Computer Vision --
  ECCV 2020}}. \bibinfo{pages}{702--717}.
\newblock


\bibitem[\protect\citeauthoryear{Zagoruyko and Komodakis}{Zagoruyko and
  Komodakis}{2016}]%
        {wide_resnet}
\bibfield{author}{\bibinfo{person}{Sergey Zagoruyko} {and}
  \bibinfo{person}{Nikos Komodakis}.} \bibinfo{year}{2016}\natexlab{}.
\newblock \showarticletitle{{Wide Residual Networks}}.
\newblock \bibinfo{journal}{\emph{arXiv preprint arXiv:1605.07146}}
  (\bibinfo{year}{2016}).
\newblock


\bibitem[\protect\citeauthoryear{Zoph and Le}{Zoph and Le}{2016}]%
        {quoc_le}
\bibfield{author}{\bibinfo{person}{Barret Zoph} {and} \bibinfo{person}{Quoc~V
  Le}.} \bibinfo{year}{2016}\natexlab{}.
\newblock \showarticletitle{{Neural Architecture Search with Reinforcement
  Learning}}.
\newblock \bibinfo{journal}{\emph{arXiv preprint arXiv:1611.01578}}
  (\bibinfo{year}{2016}).
\newblock


\bibitem[\protect\citeauthoryear{Zoph, Vasudevan, Shlens, and Le}{Zoph
  et~al\mbox{.}}{2018}]%
        {cell_1}
\bibfield{author}{\bibinfo{person}{Barret Zoph}, \bibinfo{person}{Vijay
  Vasudevan}, \bibinfo{person}{Jonathon Shlens}, {and} \bibinfo{person}{Quoc~V
  Le}.} \bibinfo{year}{2018}\natexlab{}.
\newblock \showarticletitle{{Learning Transferable Architectures for Scalable
  Image Recognition}}. In \bibinfo{booktitle}{\emph{Proceedings of the IEEE
  Conference on Computer Vision and Pattern Recognition (CVPR)}}.
  \bibinfo{pages}{8697--8710}.
\newblock


\end{thebibliography}
